\documentclass[journal,twoside,web]{ieeecolor}
\usepackage{jsen}
\usepackage{cite}
\usepackage{amsmath,amssymb,amsfonts}

\usepackage{algorithm}
\usepackage{graphicx}
\usepackage{textcomp}
\usepackage{wrapfig}
\usepackage{booktabs}
\usepackage{algpseudocode}

\def\BibTeX{{\rm B\kern-.05em{\sc i\kern-.025em b}\kern-.08em
    T\kern-.1667em\lower.7ex\hbox{E}\kern-.125emX}}
\definecolor{abstractbg}{rgb}{0.89804,0.94510,0.83137}
\begin{document}
\title{IBIS: A Hybrid Inception-BiLSTM and SVM Ensemble for Robust Doppler-based Human Activity Recognition}
\author{Alison M. Fernandes,~Hermes I. Del Monego,~Bruno S. Chang, Anelise Munaretto,\\Helder Fontes and Rui 
Campos
\thanks{A. M. Fernandes, H. I. Del Monego, B. S. Chang, and A. Munaretto are with the Universidade Tecnológica Federal do Paraná (UTFPR), Programa de Pós-Graduação em Engenharia Elétrica e Informática Industrial (CPGEI-CT), Av. Sete de Setembro, 3165, Curitiba, 80230-901, Brazil. Helder Fontes and Rui Campos are with INESC TEC, Faculdade de Engenharia, Universidade do Porto, Rua Dr. Roberto Frias, Porto, 4200-465, Porto, Portugal. Part of this work was submitted for presentation in EuCNC 2026.}%
\thanks{Corresponding author: Bruno S. Chang (e-mail: bschang@utfpr.edu.br).}%
\thanks{This work was supported by the Coordination for the Improvement of Higher Education Personnel (CAPES) under ROR identifier: 00x0ma614 for the Article Processing Charge. This work was also supported by CNPq (444629/2024-6). This work was also partially supported by Instituto de Engenharia de Sistemas e Computadores, Pesquisa e Desenvolvimento do Brasil (INESC P\&D Brasil). This research is part of the Instituto Nacional de Ciência e Tecnologia (INCT) of Intelligent Communications Networks and the Internet of Things (ICoNIoT), funded by CNPq (proc. 405940/2022-0) and CAPES (Finance Code 88887.954253/2024-00).}%
}

\IEEEtitleabstractindextext{%
\fcolorbox{abstractbg}{abstractbg}{%
\begin{minipage}{\textwidth}%

\begin{abstract}
Wi-Fi sensing is a leading technology for Human Activity Recognition (HAR), offering a non-intrusive and cost-effective solution for healthcare and smart environments. Despite its potential, existing methods struggle with domain shift issues, often failing to generalize to unseen environments due to overfitting. This paper proposes IBIS, a robust ensemble framework combining Inception-Bidirectional Long Short-Term Memory (BiLSTM) for feature extraction and Support Vector Machine (SVM) for classification of Doppler signatures. The proposed architecture specifically targets generalization capabilities. Experimental results on multiple datasets show that IBIS achieves 95.40\% accuracy, delivering a 7.58\% performance gain compared to standard architectures in cross-scenario evaluations on external datasets. The analysis confirms that IBIS effectively mitigates environmental dependency in Wi-Fi-based HAR.
\end{abstract}

\end{minipage}}}

\maketitle

\section{Introduction}
\label{sec:introduction}

\label{sec:introduction}
Wi-Fi sensing is an increasingly important technology that leverages ubiquitous Wi-Fi signals for non-intrusive Human Activity Recognition (HAR) \cite{Ahmad}. Unlike traditional methods that rely on cameras or wearable sensors, Wi-Fi sensing utilizes Channel State Information (CSI) to capture and analyze subtle changes in signal propagation caused by human movement~\cite{Yang}. This approach is not only cost-effective but also preserves user privacy, making it highly suitable for applications in healthcare, surveillance, and smart homes.

However, a key challenge in this field is overfitting, where models trained on a specific dataset fail to generalize to new environments or unseen data. Many existing solutions, particularly those based on Convolutional Neural Networks (CNNs)~\cite{chen2024} \cite{khan2021}, are effective at capturing local patterns but often struggle to learn the broader, temporal dependencies critical for robust generalization. Furthermore, raw CSI data frequently contains noise from hardware artifacts, such as phase and frequency oscillations, which can significantly hinder accurate movement interpretation and requires a rigorous sanitization process.

Recent studies have explored various approaches to enhance Wi-Fi sensing capabilities. 
Son et al.~\cite{Son} improved indoor occupancy detection by proposing a pre-processing method that uses nonlinear transformations to expand CSI data into three channel vectors, which amplified signal differences. A CNN was then applied, achieving an average accuracy of approximately 95\%. This work highlights the importance of effective signal pre-processing for boosting classification performance.

Ensemble learning has emerged as a key strategy to improve the reliability and accuracy of machine learning models. Wei et al.~\cite{Wei} used an ensemble learning approach for HAR by combining feature extraction with an SVM classifier. 

Similarly, Ye et al.~\cite{Ye} developed a system for intrusion monitoring that combines Deep Neural Networks (DNNs) with BiLSTM networks. While their approach demonstrated improved performance over traditional methods, they noted a significant weakness in detecting specific actions, such as falling movements, with an accuracy of only 62.65\%.

The SHARP algorithm, as proposed by Meneghello et al.~\cite{Meneghello2023}, provides a foundational method for HAR based on Doppler traces. Their work addressed noise and hardware interference and, after applying a neural network-based refinement, achieved an accuracy of up to 95\%. The authors also demonstrated that antenna merging could further enhance accuracy. Building upon this, Cominelli et al.~\cite{Cominelli} investigated the impact of Wi-Fi 6 features and developed a large-scale dataset for benchmarking. However, they reported difficulties in reproducing the SHARP algorithm’s results, suggesting a sensitivity to environmental variability. Their findings also indicated that SHARP's performance degrades with decreased bandwidth and a higher number of activity classes.

Despite recent advances, Wi-Fi sensing still faces limited generalization to unseen data. Prior work, such as the CNN-ABLSTM cross-subject transfer learning method by He et al.~\cite{He}, achieved 85.82\% accuracy but highlighted the need for more robust models. Moreover, most existing studies rely only on the CSI amplitude, overlooking phase information, which is highly sensitive to noise and interference.

To address these gaps, we propose IBIS, a hybrid Inception-BiLSTM architecture combined with a non-linear SVM in the post-processing stage. Unlike conventional ensemble approaches that fuse models during training, IBIS leverages the probability outputs of internal neural layers as high-level feature vectors for the SVM. Our framework operates solely on Doppler traces extracted from the CSI phase component, exploiting information typically ignored in the literature.

Preliminary versions of this work were presented at the 2025 Wireless Days conference \cite{fernandes2025wd} and at the 2026 European Conference on Networks and Communications (EuCNC) \cite{fernandes2026}, specifically within the Artificial Intelligence / Machine Learning track. The first focuses on the initial framework design and its core capabilities, while the second one shows the robustness of the proposed structure with respect to lower bandwidth scenarios. 
Expanding the introductory analysis presented in~\cite{fernandes2025wd,fernandes2026},  IBIS was evaluated on five- and eight-class activity recognition tasks and compared against two state-of-the-art baselines: the SHARP algorithm~\cite{Meneghello2023} and an ABLSTM model using raw CSI~\cite{He}. To assess robustness across environments, we also included a complementary Doppler-based dataset~\cite{Cominelli}. The proposed method achieved an average accuracy of 95.40\%, significantly outperforming all benchmark models and demonstrating strong generalization capabilities across different datasets and noise conditions.

The main contribution of this work is to enhance cross-domain generalization through a probability-space refinement using an SVM in the post-processing stage, which acts as a robust physical noise filter while relying on Doppler traces to obtain highly discriminative motion-related signals. By capitalizing on this pipeline, our method successfully extracts richer spatio-temporal information from Doppler signatures with cleaner multipath profiles, focusing strictly on the underlying dynamics of body movement.

\begin{table*}[hbt!]
\caption{Survey of Models and Features in Using CSI}
\label{tab:artigos0}
\centering
\resizebox{\textwidth}{!}{
\begin{tabular}{p{2.5cm} *{7}{c} c}
\toprule
Method / Authors & Year & Task & Model & Bandwidth & Network & Accuracy & CSI Data\\
\midrule
Son et al. \cite{Son} &  2024 & Indoor Ocupancy & CNN & 20 MHz & 802.11 &95\% &  Amplitude  \\
WiAres \cite{Wei} &  2020 & HAR & CNN+RF+SVM & - & 802.11n &98.85\% &  Temporal / Spatial Correlation  \\
SHARP \cite{Meneghello2023} &  2023 & HAR & Inception & 20/40/80 MHz & 802.11 &88.12\% &  Phase CFR  \\
Cominelli et al. \cite{Cominelli} &  2023 & HAR & Inception & 20/40/80/160 MHz & 802.11ac/ax &77.12\% &  Phase CFR  \\
Ye et al. \cite{Ye} &  2024 & HAR & DNN-ABLSTM & - & 802.11n &92.06\% &  Amplitude and Phase  \\
He et al. \cite{He} &  2025 & HAR & CNN-ABLSTM & - & - &85.82\% &  Raw CSI  \\
Widar \cite{Qian} &  2022 & Gesture Recognition & CNN+RNN & - & - &92.7\% &  Amplitude  \\
EfficientFi \cite{Jianfei} &  2022 & HAR & CNN & 40 MHz & 802.11ax &98\% &  Amplitude  \\
WiGr \cite{Xie} &  2022 & Gesture Recognition & CNN+LSTM & 40 MHz & 802.11ax &86.8\%–92.7\% &  Amplitude  \\
Wi-SensiNet \cite{Fu} &  2024 & HAR & CNN+BiLSTM & 20 MHz & 802.11 &99\% &  Amplitude  \\
IBIS* &  2026 & HAR & Inception-BiLSTM+SVM & 80/160 MHz & 802.11ac/ax &95.40\% &  Phase CFR  \\
\bottomrule
\end{tabular}
}
\begin{flushleft}
\footnotesize{Note: The symbol * represents our proposal Inception - BILSTM + SVM architecture}
\end{flushleft}

\end{table*}

It is important to highlight that our proposal achieved better results compared to the related works presented in Table \ref{tab:artigos0}. In terms of accuracy, the IBIS framework outperformed the works that use only amplitude. The main point is that when using phase information, it must be carefully handled to avoid possible classification errors caused by interference and existing offsets.

The remainder of this paper is organized as follows. Section \ref{sec:proposta} presents the definition of CSI as well as a detailed description of the datasets used in this study. Section \ref{sec:system} describes the development and implementation of the proposed algorithms, including the specific neural network architecture. The validation process, encompassing activity classification confusion matrices and Receiver Operating Characteristic (ROC) curve analysis, is presented in Section \ref{sec:avaliacao}. Finally, Section \ref{sec:conclusao} summarizes the main findings and provides concluding remarks.

\section{Channel State Information Data Description}
\label{sec:proposta}

This section presents the definition of CSI as adopted in IEEE 802.11 standard networks and also describes the datasets used in this proposal for data collection.

\subsection{Channel State Information}
CSI is a data collection system that describes how a wireless signal propagates from the transmitter to the receiver. CSI captures the signal propagation characteristics in physical environments, including the combined effects of diffraction, reflection, and scattering \cite{Son}.

Adopting the IEEE 802.11 standard, it employs advanced techniques such as Multiple-Input Multiple-Output (MIMO) and Orthogonal Frequency Division Multiplexing (OFDM) at the physical layer. These technologies aim to improve the supported data capacity and enhance channel orthogonality in environments affected by multipath propagation, which occurs when transmitted signals between the transmitter and receiver follow different paths due to phenomena like collisions and external interference.

For each transmitter/receiver antenna pair, CSI provides detailed information about the phase and amplitude attenuation experienced by the signal through multiple paths for each subcarrier. Specifically, the CSI for each subcarrier $k$ between a transmitter with $N_t$ antennas and a receiver with $N_r$ antennas is represented by the matrix $\mathbf{H}(k) \in \mathbb{C}$ \cite{Son}:
\begin{equation}
    \mathbf{H}(k) = 
\begin{bmatrix}
h_{11}(k) & \cdots & h_{1N_t}(k) \\
h_{21}(k) & \cdots & h_{2N_t}(k) \\
\vdots & \ddots & \vdots \\
h_{N_r1}(k) & \cdots & h_{N_rN_t}(k),
\end{bmatrix}, k \in [1,K] 
\end{equation}
where $K$ represents the number of subcarriers used, and $h_{rc}(k)$ denotes the complex channel gain from the $c{\text{-th}}$ transmitter antenna to the $r{\text{-th}}$ receiver antenna. Each element $h_{rc}(K)$ in the CSI matrix $\mathbf{H}(k)$ can be decomposed into amplitude and phase components, as described by the following:
\begin{equation}
h_{rc}(k) = |h_{rc}(k)| \cdot \exp\left(j \theta_{rc}(k)\right),
\end{equation}
where $\left| h_{rc}(k) \right|$ is defined as the amplitude component, and $\theta_{rc}(k)$ is the phase component of $h_{rc}(k)$.

Since CSI contains both amplitude and phase information for each subcarrier and for each pair of transmit and receive antennas, it provides high-resolution data compared to simple measurements such as RSS (Received Signal Strength). In this way, CSI data can be considered as a form of "imaging" of the signal propagation environment.

Changes in the environment, such as the presence or movement of people, directly affect Wi-Fi signal propagation. These alterations in the signal path also lead to variations in CSI. Since different people create distinct patterns, it is possible to estimate the number of individuals present by learning these patterns—this is the key concept behind people counting systems based on CSI-extracted data \cite{Son}.

\subsection{Dataset A}

One of the datasets used for this research is based on the SHARP study~\cite{Meneghello2023} and consists of CSI data collected from three volunteers (one male, two females) performing eight activities in various indoor environments. These activities: empty, sitting, stand up, walking, running, jumping, and arm gym, were each recorded for 120 seconds. The data was gathered in a bedroom, living room, and university laboratory during two periods: April–December 2020 and January 2022. A Nexmon CSI Extraction Tool and an Asus RT-AC86U Wi-Fi router were used to extract the CSI from Wi-Fi 2.4 GHz preamble headers.

CSI is a data framework that describes how a wireless signal travels from a transmitter to a receiver. It provides high-resolution data on the physical propagation properties of a signal, including the effects of diffraction, reflection, and scattering~\cite{Son}. Because CSI captures both amplitude and phase information, it is sensitive to environmental changes, such as the movement of people. 

The dataset includes seven distinct scenarios (S1-S7):

\begin{itemize}
    \item S1–S3: A bedroom with a central bookcase. S1 was used for model training, S2 for a generalization scenario, and S3 was measured on different days than S1;
    \item S4–S5: Activities performed alternately within the bedroom;
    \item S6: Data collected in a living room on different days;
    \item S7: A laboratory with numerous desks, computers, and monitors, measured over several days.
\end{itemize}
   
Scenario S7 was chosen as the benchmark for this study's model evaluation because it presented the greatest challenge. According to the SHARP authors, the high density of obstacles and reflective surfaces in this environment significantly impacts CSI signal quality and classification performance.

\subsection{Dataset B}
For application in the experiment, we considered using a second dataset, developed by Cominelli et al. \cite{Cominelli}.  
Although originally designed for IEEE 802.11ax (Wi-Fi 6), thereby expanding the application scope. The main characteristics of this dataset are summarized below.

This dataset was developed using four Asus RT-AX86U routers interconnected to form a local-area network (LAN) and controlled by a host notebook. One of the routers is configured to generate dummy Wi-Fi traffic at a constant rate through feature injection for AX-CSI, while the other three extract the corresponding CSI. According to the authors, due to the use of IEEE 802.11ax, data grows very rapidly, making it necessary to store the recordings in PCAP files.

Data collection was performed across seven different scenarios:

\begin{itemize}
    \item S1 to S5: Conducted in the same physical environment, where neither the receivers nor the transmitter were moved during the entire duration of the experiments;
    \item S6: An office environment containing chairs and desks;
    \item S7: A large hall, representing an open environment without obstacles.
\end{itemize}

Twelve distinct activities were recorded:
Walk, Run, Jump, Sitting, Empty room, Standing, Wave hands, Clapping, Lay down, Wiping, Squat, Stretching
with a sampling rate of 80 samples per second.

For our experiment, scenario S2 was selected.
This environment consists of a laboratory with reflective surfaces, which makes prediction more difficult.

Due to the accuracy challenges encountered with the original model, our intention is to apply the IBIS framework in different environments and scenarios in order to evaluate its behavior and robustness.

\section{System Modeling}
\label{sec:system}

In this section, the main definitions related to the IBIS framework are presented, including the architectural composition, as well as the derived pseudocode outlining the main steps of the framework operation. The code used to conduct these experiments will be made available in a GitHub repository following the publication of this paper.

\subsection{Decision-Making Process and Algorithm Steps}

\begin{figure*}[ht!]
    \begin{center}
        \includegraphics[width=0.75\textwidth]{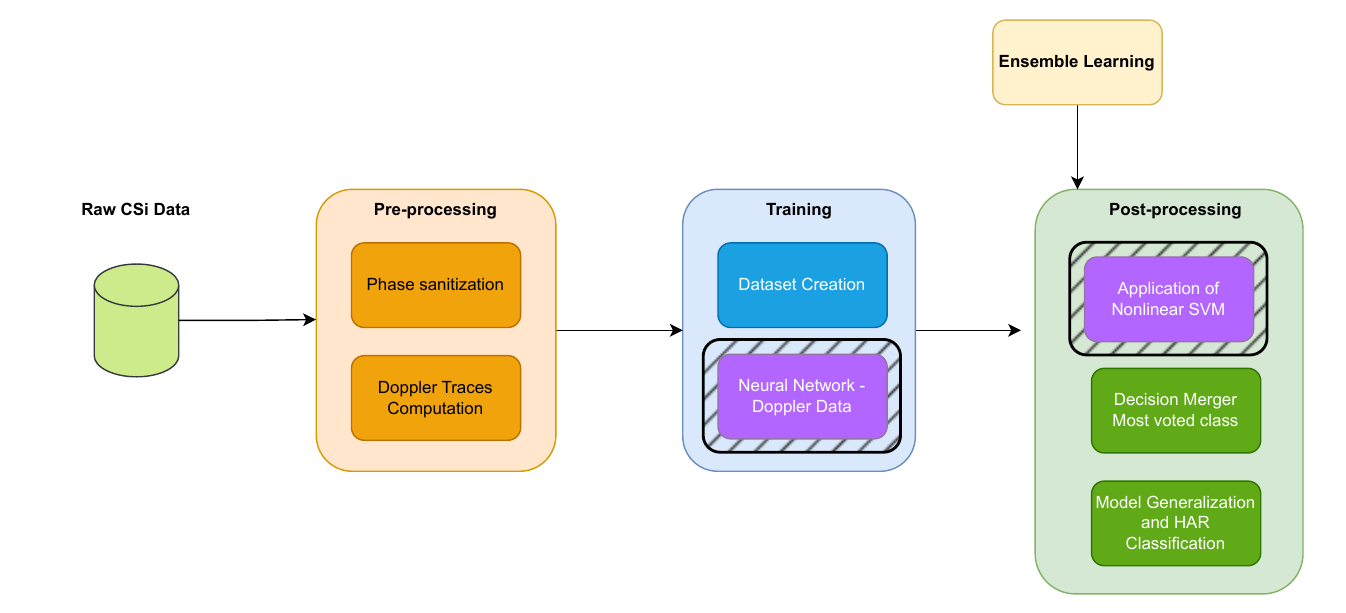}
        \caption{The sequential process diagram illustrates the IBIS steps: collection of raw CSI data, passing through sanitization, neural network training, and Ensemble Learning post-Processing.}
        \label{fig:sharp_}
    \end{center}    
\end{figure*}

To understand the purpose of our work, it is essential to first comprehend the SHARP method. After the CSI data is extracted from the Wi-Fi 802.11 ac/ax packet preambles, it undergoes three key stages, as shown in Figure ~\ref{fig:sharp_}:

\subsubsection{Pre-processing}
This stage sanitizes the CSI data by eliminating noise, interference, and phase ambiguity. A reference signal path is used to remove phase offsets, recovering a clean signal that reveals the Doppler traces associated with movement.

\subsubsection{Doppler Traces Extraction}

Initially, during the sanitization process, artifacts and interference are removed to facilitate the extraction of Doppler traces. For this purpose, a sliding window segmentation is applied, where the filtered and normalized CSI matrix $X \in \mathbb{C}^{M \times K}$ (where $M$ is the total number of packets and $K$ is the number of selected subcarriers) is divided into small temporal windows of fixed length. After this step, a time-shifting factor is applied to increase the temporal resolution of the Doppler profiles. Subsequently, a Hann window \cite{Haitao} is applied along the time axis to attenuate spectral leakage. Finally, the Fast Fourier Transform (FFT) and the Power Spectral Density (PSD) are calculated.Following the Doppler extraction process, the scenarios are selected for training, validation, and testing. To verify the generalization capability and ensure that the model does not suffer from data leakage, the training and testing scenarios are strictly distinct. 

\subsubsection{Classification}
The sanitized Doppler traces are fed into an Inception neural network. To improve generalization beyond the original architecture's spatial feature learning capabilities, the network was enhanced with a BiLSTM layer, creating a hybrid Inception–BiLSTM model. This modification enables the architecture to capture both spatial and temporal dependencies, significantly boosting performance in human activity recognition. At this stage, it is possible to select the number of antennas used in the training process. The more antennas that are selected, up to the limit of four,the more distinctive information becomes available, consequently improving the neural network’s recognition performance.

To understand how the IBIS operates, it is important to recognize the characteristics and advantages introduced by this new configuration. Unlike the original Inception network, additional layers are incorporated, providing enhanced data interpretation and improving the model’s ability to extract discriminative features.

Figure~\ref{fig:neural_v3} illustrates the integrated architecture of these two neural structures. While some layers like Normalization, Reshape, and Dropout are omitted for visualization, they are described in the text. Table~\ref{tab:model_architecture_compact} provides the input and output dimensions and parameter counts for each block. The shaded and hatched purple areas in the figure highlight the novel components of our proposal.

Unlike traditional sequential architectures, a key novelty in the proposed system modeling lies in the tight coupling between the temporal attention-driven features and the robust margin maximization of the SVM ensemble through the probability outputs. This specific architectural enhancement explicitly targets feature generalization, preventing the classifier from overfitting to environment-specific Doppler artifacts.

Our approach enhances the standard Inception network, which typically only includes activation and convolution layers for learning local features. We added a MaxPooling layer and Batch Normalization to improve convergence and filter relevant features. Furthermore, a BiLSTM network was incorporated to enable the model to learn temporal patterns and motion dynamics from Doppler shifts.

\textbf{Inception Network:} This module is configured to extract spatial features from the Doppler trace data. A description of each block and its function is as follows:

\begin{itemize}
    \item Input: The entry point for the Doppler trace data;
    \item Normalization: Normalizes input data by subtracting the mean and dividing by the standard deviation to stabilize training and improve convergence speed;
    \item Conv2D (Conv Block): This block combines Conv2D, Batch Normalization, and an Activation function. The Conv2D layer uses filters to extract local patterns like edges and textures;
    \item Batch Normalization: Normalizes activations to accelerate training and regularize the model;
    \item Activation: Introduces non-linearity, allowing the model to learn complex representations; 
    \item Max Pooling: Reduces spatial resolution and complexity to focus on the most relevant features;
    \item Concatenate: Merges activation maps from different paths, maintaining important information while reducing dimensionality;
    \item Dropout: Randomly deactivates neurons during training to reduce overfitting and improve generalization;
    \item Reshape: Transforms the output tensor into a sequence format, preparing it for temporal processing by the BiLSTM layer.
\end{itemize}

The combination of Conv2D, Batch Normalization, and Activation is repeated nine times within the Inception module, a core characteristic that increases depth and improves generalization by learning more abstract features.

\textbf{BiLSTM Network}: The second part of the architecture is the BiLSTM network, which was included to capture crucial temporal patterns and dependencies from the Doppler traces. These traces are time-series signals that reflect movement and channel variations, often exhibiting complex patterns that the network must recognize.
 
The BiLSTM network's components are described as follows: 

\begin{itemize}
    \item BiLSTM + Attention: This layer processes input sequences in both forward and backward directions to capture complex temporal relationships. Although BiLSTM networks are highly efficient at processing entire sequences, they can suffer from information degradation or fading temporal gradients when handling extremely long sequences. To mitigate this limitation, an attention mechanism is integrated. Instead of treating all sequence time steps equally, the attention layer dynamically assigns weights, enabling the model to emphasize and recognize the most critical temporal features. The primary benefit of combining the attention mechanism with the BiLSTM architecture is that it provides optimal inputs for the subsequent SVM stage. Since SVM performance relies heavily on highly separable data, raw or noisy features would severely degrade the boundary separation process. The attention mechanism condenses the sequence into a high-quality context vector that retains only the most significant points for the SVM, thereby maximizing the overall generalization capability.
    \item Global Average Pooling: This layer reduces the dimensionality of the feature space in preparation for the subsequent processing stages. In contrast to traditional architectures that utilize a Flatten layer—which often leads to overfitting due to the generation of excessively large feature vectors—Global Average Pooling computes the average value of each feature map. By compressing this dimensionality, the model filters out noise and redundant data while preserving only the essential spatial and temporal characteristics, resulting in a cleaner dataset for the classifier.
    \item Dense Layer: A fully connected layer that acts as a dimensionality adjustment tool for the SVM stage. It ensures that the network outputs the precise number of dimensions required by configuring the exact number of output units. This process transforms the deep features into a compact, high-quality representation vector, where the final number of units depends directly on the number of target output classes(Table \ref{tab:model_architecture_compact}).

\end{itemize}

\begin{figure*}[ht!]
    \begin{center}
      \includegraphics[width=0.75\textwidth]{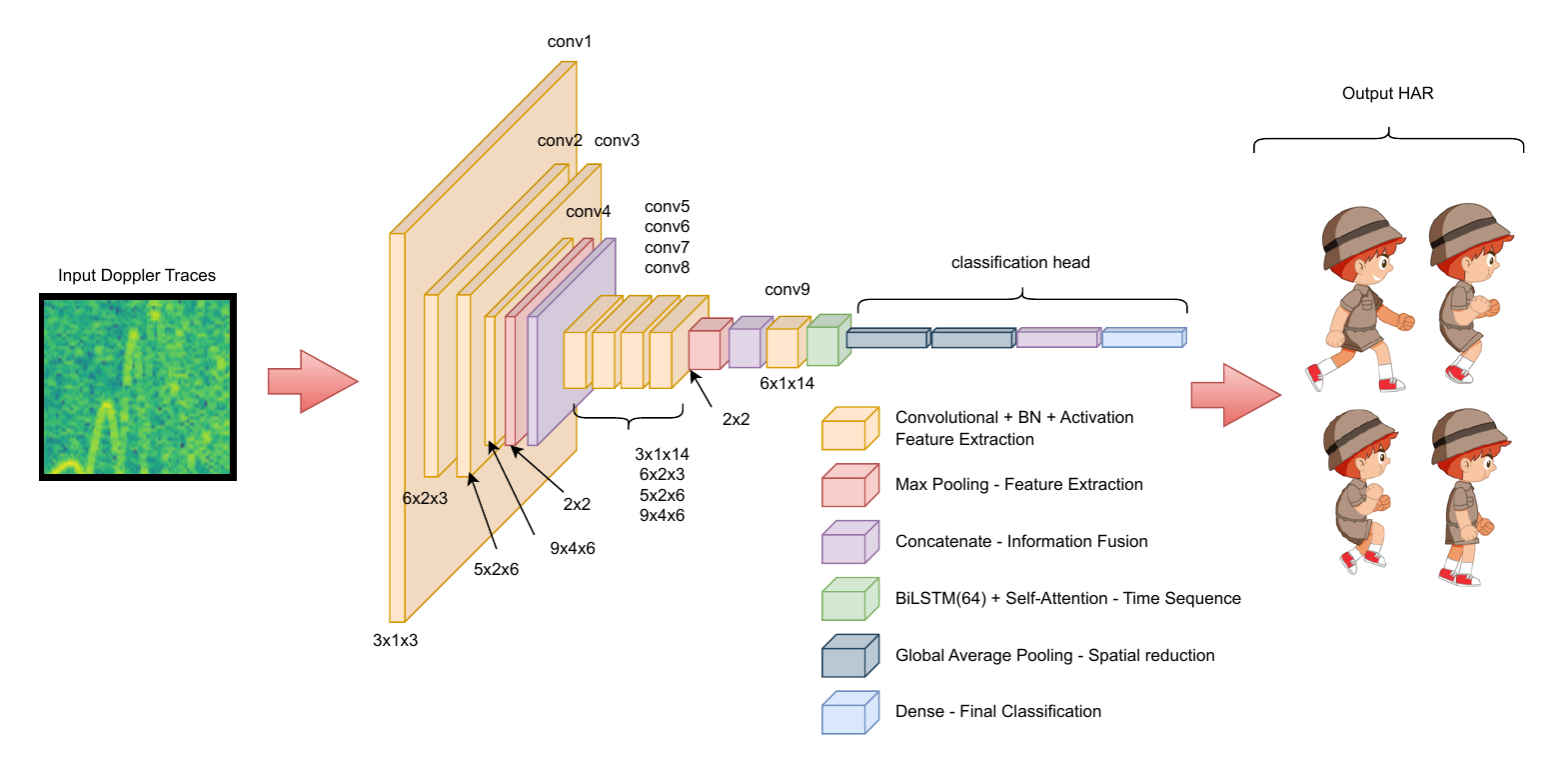}
        \caption{Architecture of the Inception-BiLSTM hybrid neural network for Human Activity Recognition using Doppler traces, showing all convolutional and attention layers.}
        \label{fig:neural_v3}
    \end{center}    
\end{figure*}

\begin{table}[ht!]
\caption{Inception-BiLSTM Architecture (Compact Notation)}
\label{tab:model_architecture_compact}
\begin{center}
\begin{tabular}{p{3.8cm}|p{2.7cm}|p{1.0cm}}\toprule
\textbf{Block / Layer} & \textbf{Output Shape} & \textbf{Parameters} \\ \midrule
Input Layer & (32, 32, 3) & - \\
Normalization & (32, 32, 3) & mean, variance \\ \midrule

Conv2D(3×1×3) & (32, 32, 3) & 3 \\
BatchNormalization & (32, 32, 3) & 12 \\
Activation (swish) & (32, 32, 3) & - \\
Conv2D(6×2×3) & (31, 31, 6) & 78 \\
BatchNormalization & (31, 31, 6) & 24 \\
Activation (swish) & (31, 31, 6) & - \\
Conv2D(5×2×6) & (30, 30, 5) & 125 \\
BatchNormalization & (30, 30, 5) & 20 \\
Conv2D(9×4×6) & (28, 28, 9) & 873 \\
BatchNormalization & (28, 28, 9) & 36 \\
Activation (swish) & (30×30×5) / (28×28×9) & - \\
MaxPooling2D(2×2) & (15, 15, 5) & - \\
Concatenate & (15, 15, 14) & - \\ 

Conv2D(3×1×14) & (15, 15, 3) & 45 \\
BatchNormalization & (15, 15, 3) & 12 \\
Activation (swish) & (15, 15, 3) & - \\
Conv2D(6×2×3) & (14, 14, 6) & 78 \\
BatchNormalization & (14, 14, 6) & 24 \\
Conv2D(5×2×6) & (13, 13, 5) & 125 \\
BatchNormalization & (13, 13, 5) & 20 \\
Conv2D(9×4×6) & (11, 11, 9) & 873 \\
BatchNormalization & (11, 11, 9) & 36 \\
Activation (swish) & (13×13×5) / (11×11×9) & - \\
MaxPooling2D(2×2) & (6, 6, 5) & - \\
Concatenate & (6, 6, 14) & - \\
Dropout(0.5) & (6, 6, 14) & - \\

Conv2D(6×1×14) & (6, 6, 6) & 84 \\
BatchNormalization & (6, 6, 6) & 24 \\
Activation (swish) & (6, 6, 6) & - \\
Reshape & (36, 6) & - \\ \midrule

Bidirectional LSTM (64 units) & (36, 128) & 45568 \\
Activation (tanh) & (36, 128) & - \\
Attention & (36, 128) & 16640 \\
GlobalAvgPooling1D (RNN) & (128,) & - \\
GlobalAvgPooling1D (Attention) & (128,) & - \\
Concatenate & (256,) & - \\
Dropout(0.5) & (256,) & - \\
Dense (5 classes) & (5,) & 1285 \\
\bottomrule
\end{tabular}
\end{center}
\raggedright
\footnotesize{Note: `Conv2D(out×kernel×in)` notation refers to output channels, filter size (square or rectangular kernel), and input channels, respectively. Shapes may vary slightly depending on padding/stride.}
\end{table}

\subsection{Post-Processing Support Vector Machine}

The hybrid Inception-BiLSTM neural network is effective at detecting and classifying both global patterns and temporal sequences. However, for accurate classification of CSI-derived data, this architecture alone is insufficient. To address this, an SVM was integrated into the post-processing stage as an ensemble learning strategy. Unlike common approaches where SVM is the primary classifier, this work uses it to refine the final classification results and improve generalization.
The SVM's ability to handle outliers and its use of decision margins contribute to better model generalization, which is particularly useful for Doppler trace data that is prone to noise and distortions. This allows the SVM to refine the final classification results, especially for samples that might be misclassified by the neural network.

A key advantage of SVM lies in the flexibility of its kernel functions. The Radial Basis Function (RBF) kernel, as well as the Sigmoid and Polynomial kernels \cite{kolodiazhnyi}\cite{ye2012}\cite{Arti}, are available options that can be selected for classification. In our framework, during the training stage, the SVM is trained using the probability outputs generated by the neural network. Through a grid-search optimization process, the most suitable kernel is determined for the underlying class distribution.

The penalty parameter C is also tuned to achieve an optimal balance between training accuracy and model generalization. Furthermore, class balancing was applied to the SVM configuration to ensure that all classes receive equal weighting, preventing the classifier from becoming biased toward majority classes and resulting in a fairer and more robust overall prediction performance.

\begin{figure}[ht]
    \begin{center}
      \includegraphics[width=0.5\textwidth]{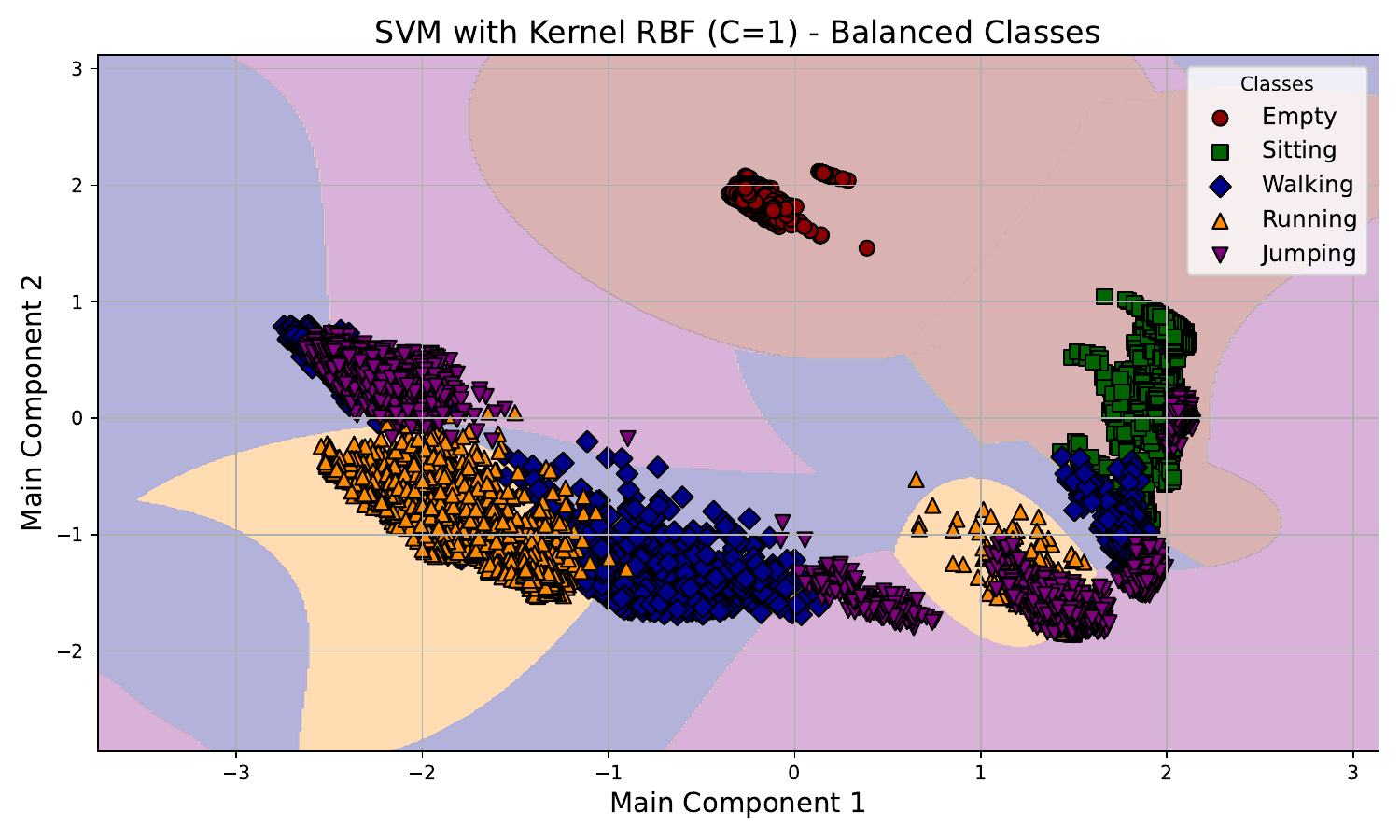}
        \caption{Representation of the SVM algorithm applied to five movements with non-linear Data, applying the PCA method for dimensionality reduction.}
        \label{fig:svm_class}
    \end{center}    
\end{figure}

Figure~\ref{fig:svm_class} illustrates the SVM model's application, showing five distinct classes separated by well-defined boundaries. Proper class segmentation is essential for understanding the unique features of each class, leading to improved classification accuracy.

Regarding the hyperparameters, a learning rate of 0.001 was used for the hybrid network, along with an early stopping callback function to achieve the highest possible accuracy. At the classification stage, the SVM was also trained using a Grid Search, where three parameters were varied to determine the optimal training configuration. Initially, three kernel functions were considered: RBF, sigmoid, and polynomial, along with the trade-off parameter $C$ and gamma ($\gamma$), both ranging from 0.01 to 1, which are responsible for defining the decision margins. The RBF kernel was ultimately selected by the Grid Search, representing the non-linear characteristics of the data.

The main reason for incorporating an SVM in the post-processing stage is that the Softmax layer of a neural network tends to force a probabilistic decision, even in highly ambiguous classes, and is sensitive to small variations in previously unseen environments. In contrast, the SVM aims to maximize the separation margin using high-level feature vectors, receiving a cleaner representation of class boundaries. This makes the inclusion of an SVM particularly suitable for noisy Doppler traces, where decision boundaries are not well defined. In this context, the combination of the BiLSTM module with dropout regularization, followed by the SVM in the post-processing step, contributes significantly to improving the model’s generalization capability.

\subsection{Pseudocode}

The pseudocode description presented in algorithm ~\ref{alg:ibis_pseudocode} concisely illustrates the application of the IBIS framework and the main processes involved. 

It is worth mentioning that the code containing the modifications to the SHARP algorithm, as well as all stages of the IBIS implementation, will be made public after the publication of this proposal.

\begin{algorithm}[t]
\caption{IBIS framework}
\label{alg:ibis_pseudocode}
\begin{algorithmic}[1]

\State \textbf{Preprocessing:}
\State \hspace{0.5cm} Apply LASSO-based sanitization and remove offset
\State \hspace{0.5cm} Extract Doppler traces and select subcarriers

\State \textbf{Input:} Dataset $X$, labels $Y$, batch size $32$, learning rate $0.001$, epochs $100$

\State \hspace{0.5cm} Normalize features: $X_{\text{norm}} = \text{normalize}(X)$
\State \hspace{0.5cm} Split dataset: $(X_{train}, Y_{train}, X_{test}, Y_{test}) \gets \text{split}(X_{\text{norm}}, Y, 0.6)$

\State \textbf{Inception Feature Extraction:}
\State \hspace{0.5cm} Apply parallel convolutions with multiple kernel sizes
\State \hspace{0.5cm} Concatenate outputs to obtain $F_{\text{inception}}$

\State \textbf{BiLSTM Temporal Modeling:}
\State \hspace{0.5cm} Feed $F_{\text{inception}}$ into BiLSTM layers
\State \hspace{0.5cm} Extract temporal representation $F_{\text{BiLSTM}}$

\State \textbf{Softmax Classification:}
\State \hspace{0.5cm} Compute posterior probabilities: 
\State \hspace{1.0cm} $P(Y|X)=\text{Softmax}(W_s F_{\text{BiLSTM}} + b_s)$
\State \hspace{0.5cm} Perform Grid Search to select kernel type, $C$, and $\gamma$

\State \textbf{SVM Training (Postprocessing):}
\State \hspace{0.5cm} Train SVM using $(P(Y|X_{train}), Y_{train})$
\State \hspace{0.5cm} Available kernels:
\State \hspace{1.0cm} \textbf{RBF:} $K(x,x')=\exp(-\gamma\|x-x'\|^2)$
\State \hspace{1.0cm} \textbf{Sigmoid:} $K(x,x')=\tanh(\gamma\, x^\top x' + r)$
\State \hspace{1.0cm} \textbf{Polynomial:} $K(x,x')=(\gamma\, x^\top x' + r)^d$

\State \textbf{Prediction:}
\State \hspace{0.5cm} Compute $P(Y|X_{test})$ using the Inception--BiLSTM--Softmax network
\State \hspace{0.5cm} Predict: $Y_{pred} = \text{SVM.predict}(P(Y|X_{test}))$

\State \textbf{Evaluation:}
\State \hspace{0.5cm} Compute Accuracy, Precision, Recall, and F1-score

\State \textbf{Output:} Final predictions $Y_{pred}$ and evaluation metrics

\end{algorithmic}
\end{algorithm}

\section{Performance Evaluation}
\label{sec:avaliacao}

This section evaluates the IBIS framework using generalization data from Scenario S7 for Dataset A and from Scenario S2 for Dataset B, as previously described. It is important to note that this validation used an 80 MHz bandwidth (IEEE 802.11ac) for Scenario S7 and a 160 MHz bandwidth (IEEE 802.11ax) for Scenario S2, with the selection of four antennas to maximize the amount of information obtained during training. This scenario, a laboratory room filled with desks, chairs, and monitors, was chosen as a benchmark due to its challenging nature. The numerous obstacles in this setup introduce complex signal distortions like multipath propagation, reflection, and scattering, which make the task of activity recognition particularly difficult for any classifier. We will analyze the model's performance through its learning curves, class-wise confusion matrices, and the computational cost of the hybrid architecture.

Regarding the data split, the dataset is partitioned 60\%/20\%/20\% for training, validation, and testing. Critically, these partitions are drawn from physically distinct scenarios: training uses scenario S1, while validation and testing use scenario S7. Because the sliding-window segmentation is applied independently within each scenario and S1 and S7 share no recordings, no window can appear in both the training and test sets, eliminating train–test leakage from window overlap.

To avoid decision bias in the evaluated algorithms, ten simulations were performed for each experiment, and the results provided herein represent the average across the ten simulations.

Table \ref{tab:params_comparison_inception} presents all the training hyperparameters, including batch size, learning rate, number of epochs, and the optimizers involved in the simulation. Regarding the post-processing phase, a grid search framework was utilized to vary parameters such as kernel influence, gamma, and kernel type. These SVM parameters can be visualized in Table \ref{tab:tradeoff_gamma}. In this regard, the RBF kernel was selected across the multiple simulation iterations executed during the Grid Search process. This choice is primarily due to the nature of the data, as the RBF kernel is highly effective for processing non-linearly separable datasets.

Another important aspect that deserves highlight is that a low $C$ value imposes a smaller penalty on misclassification errors, allowing the model to prioritize a wider and smoother margin, thus increasing error tolerance. Conversely, a high $C$ value imposes stricter penalties on classification errors, creating more complex decision regions that may lead to overfitting.

Regarding the regularization parameter $\gamma$, a low value represents a broader radius of influence, yielding a behavior closer to a linear classification. In contrast, higher $\gamma$ values restrict the radius of influence, making the decision boundary tighter and more rigid around the analyzed data points. Therefore, considering a balanced trade-off between both $C$ and $\gamma$, the configuration of $C=1$ and $\gamma=1$ allowed the model to achieve an optimal equilibrium between these learning extremes. This setting provided sufficient flexibility to capture relevant patterns without overcomplicating the boundary, thereby ensuring a high accuracy rate.

\begin{table}[hbt!]
\caption{Hyperparameters and Experimental Settings for the Inception-BiLSTM Network}
\label{tab:params_comparison_inception}
\centering
\small
\centering
\begin{tabular}
{p{3.2cm}p{4.6cm}}
\toprule
\textbf{Parameter} & \textbf{Inception-BiLSTM Value / Setting} \\
\midrule
Architecture & Inception-BiLSTM Hybrid \\
Input Channels & 540 Channels \\
Optimizer & Adam \\
Learning Rate & $1 \times 10^{-4}$ \\
Loss Function & Sparse Categorical Crossentropy \\
Number of Max Epochs & 25 \\
Early Stopping & Validation Loss (Patience = 3) \\
Model Checkpoint Monitor & Sparse Categorical Accuracy \\
Attention Mechanism & Parallel Self-Attention + GAP + GMP \\
Classification Head & Dense Layer ($256 \times 10$) \\
Output Classes & 8 \\
\bottomrule
\end{tabular}

\vspace{2pt}
\begin{flushleft} 
\footnotesize{Note: Table with the hyperparameters used to train the proposed framework, as well as information about settings and classes}
\end{flushleft}
\end{table}

\begin{table}[hbt!]
\caption{SVM classification accuracy across different hyperparameter settings ($C$ and $\gamma$) for each bandwidth to Dataset A. Optimal performance is obtained with $C=1$ and $\gamma=1$}
\label{tab:tradeoff_gamma}
\begin{center}
\begin{tabular}{p{1.8cm}p{1.8cm}p{1.8cm}p{1.8cm}}
\toprule
$C$ & $\gamma = 0.01$ (\%) & $\gamma = 0.1$ (\%) & $\gamma = 1$ (\%) \\
\midrule
0.01 & 88.49 & 80.52 & 91.41 \\
0.1  & 90.69 & 72.90 & 91.80 \\
1    & 83.46 & 80.06 & 95.40 \\
\bottomrule
\end{tabular}
\end{center}
\footnotesize{Note: $C$ – Regularization, $\gamma$ – Kernel influence}
\end{table}

\subsection{Confusion Matrix for 5 Movements}

\begin{figure}[ht!]
    \centering
    \begin{minipage}[b]{0.24\textwidth}
        \centering
        \includegraphics[width=\textwidth]{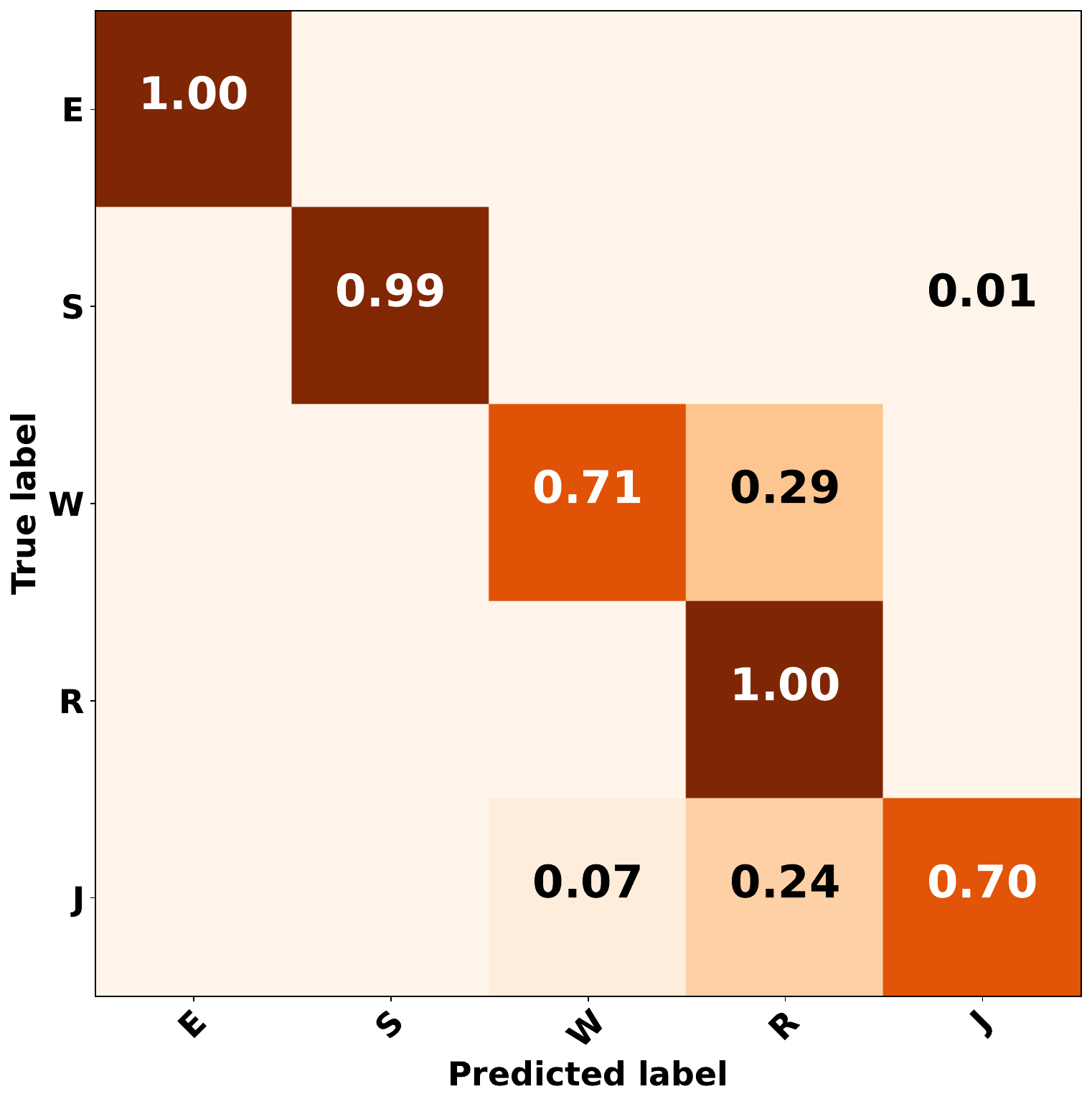}
        \text{(a) Inception - 88.16\% }
        \label{fig:conf_5_old}
    \end{minipage}
    \hfill
    \begin{minipage}[b]{0.24\textwidth}
        \centering
        \includegraphics[width=\textwidth]{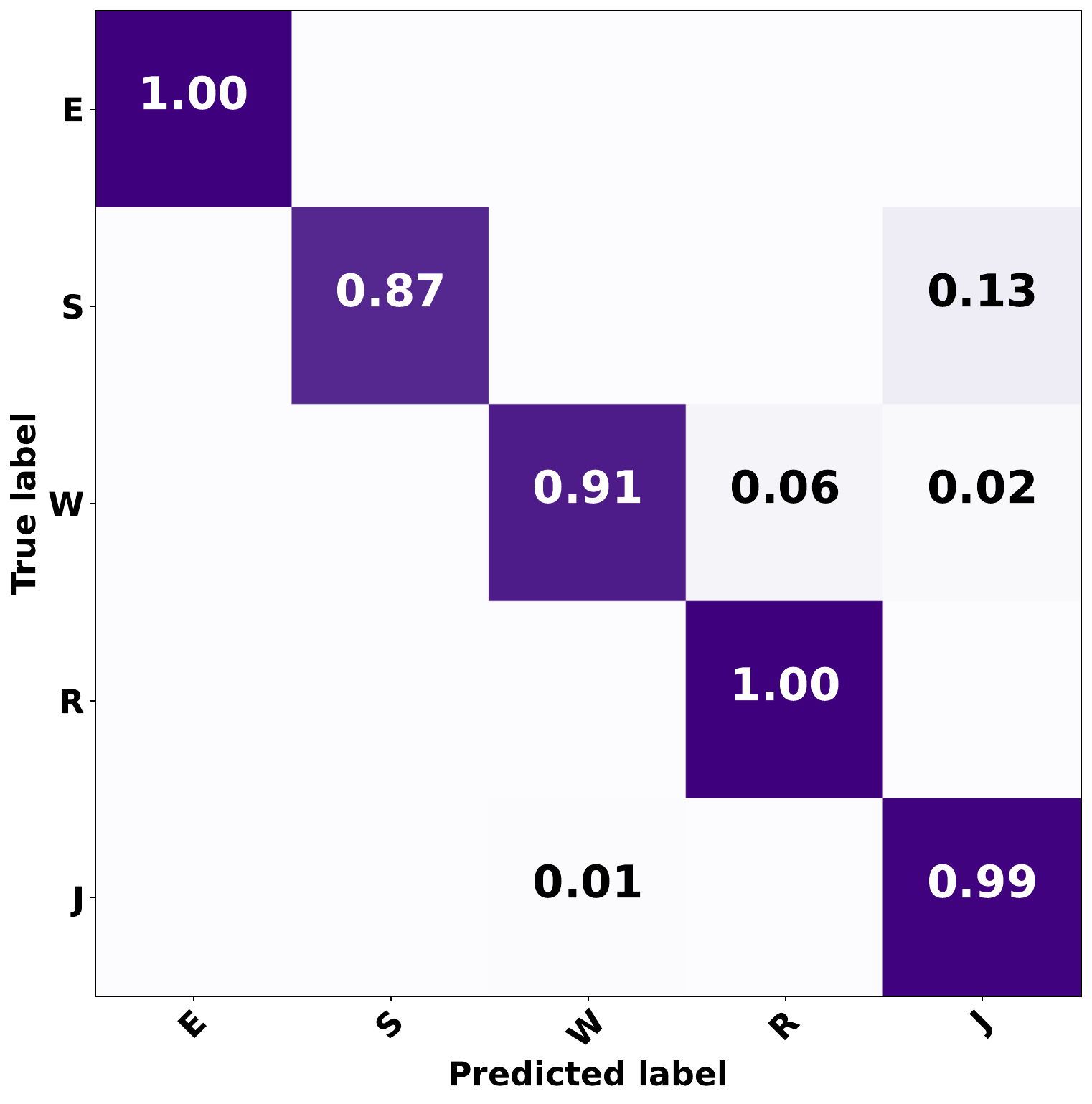}
        \text{(b) IBIS - 95.40\%}
        \label{fig:conf_5}
    \end{minipage}
   \hfill
      \footnotesize \textit{Note: E - Empty, S - Sitting, W - Walking, R - Running, and J - Jumping}
    \caption{Comparative analysis of the confusion matrix for five movements: original Inception versus IBIS}
    \label{fig:conf_gen5}
\end{figure}

Figure \ref{fig:conf_gen5} compares the confusion matrices for five movements, showing the results from the Inception-only network against the proposed hybrid scheme. As seen in Figure \ref{fig:conf_gen5}a, the Inception network struggled with precision, particularly for the walking movement, which was frequently confused with running, achieving only 71\% accuracy. Jumping was also often misclassified as running, with an accuracy of just 70\%. This lack of precision is attributed to the similarity of Doppler patterns between these movements and the fact that the Inception network was not designed to handle temporal dependencies crucial for accurate CSI data classification.

Conversely, Figure \ref{fig:conf_gen5}b shows the results obtained after implementing our IBIS framework, which achieved substantially better performance with only minor misclassifications. A more balanced distribution of accuracy across the classes is also observed, reaching up to 99\% in categories where the original model previously exhibited confusion. Overall, this configuration yields a 7.58\% performance gain compared to the previous simulation.

This improvement is primarily attributed to the BiLSTM component’s ability to capture temporal patterns from the Doppler data. In addition, the SVM, through the selection of an appropriate kernel, proper tuning of the parameter $C$, and class balancing, further refined the classification boundaries, resulting in enhanced overall performance. Both confusion matrices shown correspond to the post-antenna fusion stage, where classification decisions are merged using majority voting.

\subsection{ROC Curve for 5 Movements}

\begin{figure*}[ht!]
    \centering
    \begin{minipage}[b]{0.48\textwidth}
        \centering
        \includegraphics[width=\textwidth]{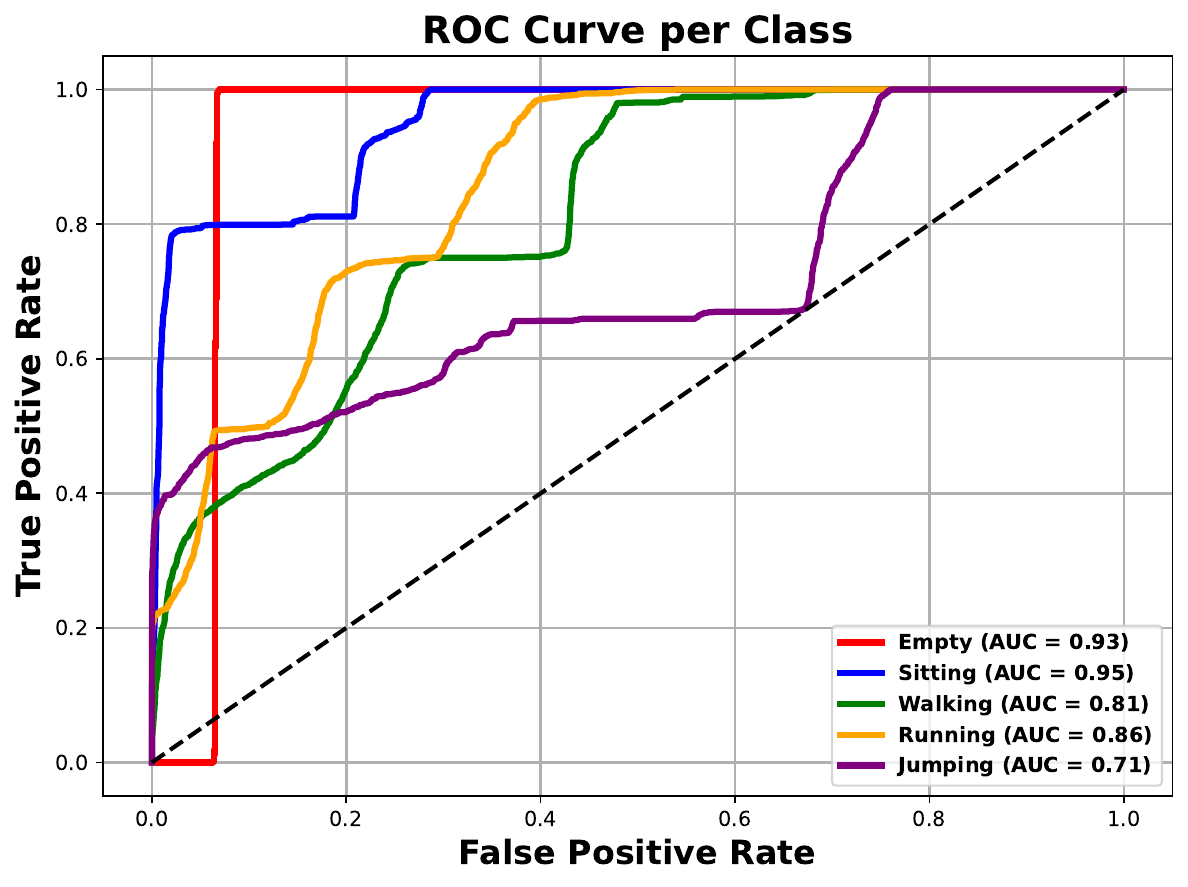}
        \text{(a) Inception}
        \label{fig:roc_old_5}
    \end{minipage}
    \hfill
    \begin{minipage}[b]{0.48\textwidth}
        \centering
        \includegraphics[width=\textwidth]{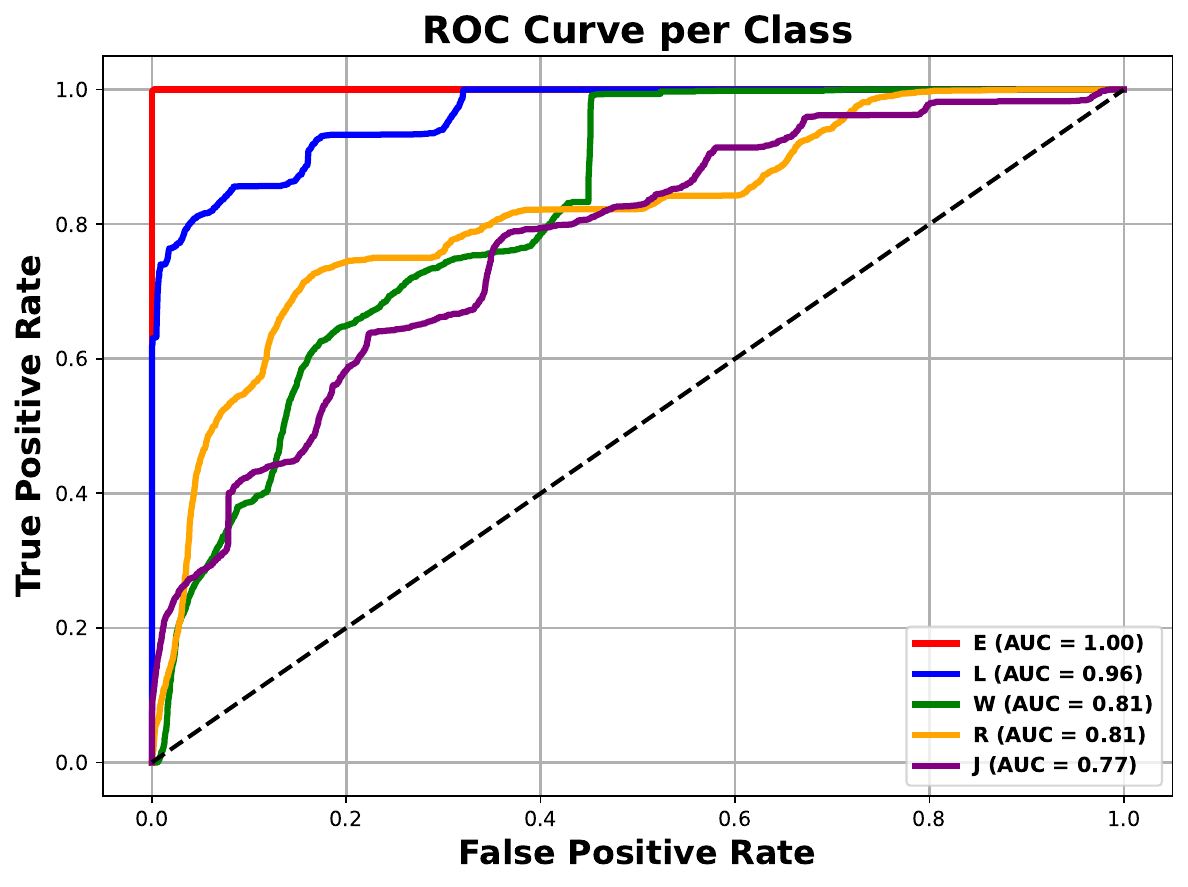}
        \text{(b) IBIS}
        \label{fig:roc_5}
    \end{minipage}
    \caption{ROC curve analysis for five movements: comparing the original Inception with IBIS}
    \label{fig:roc_gen5}
\end{figure*}

Figure \ref{fig:roc_gen5} displays the learning  ROC curve for all five classes, with a dotted diagonal line indicating the classification threshold. This curve was generated using generalization test data without antenna fusion, allowing for a direct evaluation of the algorithm's generalization capability.

As shown in Figure \ref{fig:roc_gen5}a, the original model performed poorly for the walking, running, and jumping classes, which exhibited significant fluctuations in accuracy. In contrast, Figure \ref{fig:roc_gen5}b shows that, with our proposed approach, the accuracy values increased and became more stable, with no apparent oscillations, even without antenna fusion, except for the Walking class. At this stage, we are evaluating only the effect of applying the SVM after training the neural network using the softmax probabilities as input. In other words, this result reflects exclusively the effectiveness of the SVM refinement, excluding the majority-voting step from antenna fusion, which is applied afterward. However, when considering the class distribution, the results are more uniform. This highlights the effectiveness of incorporating ensemble learning in this challenging scenario.

\subsection{Confusion Matrix for 8 Movements}

\begin{figure}[ht!]
    \centering
    \begin{minipage}[b]{0.24\textwidth}
        \centering
        \includegraphics[width=\textwidth]{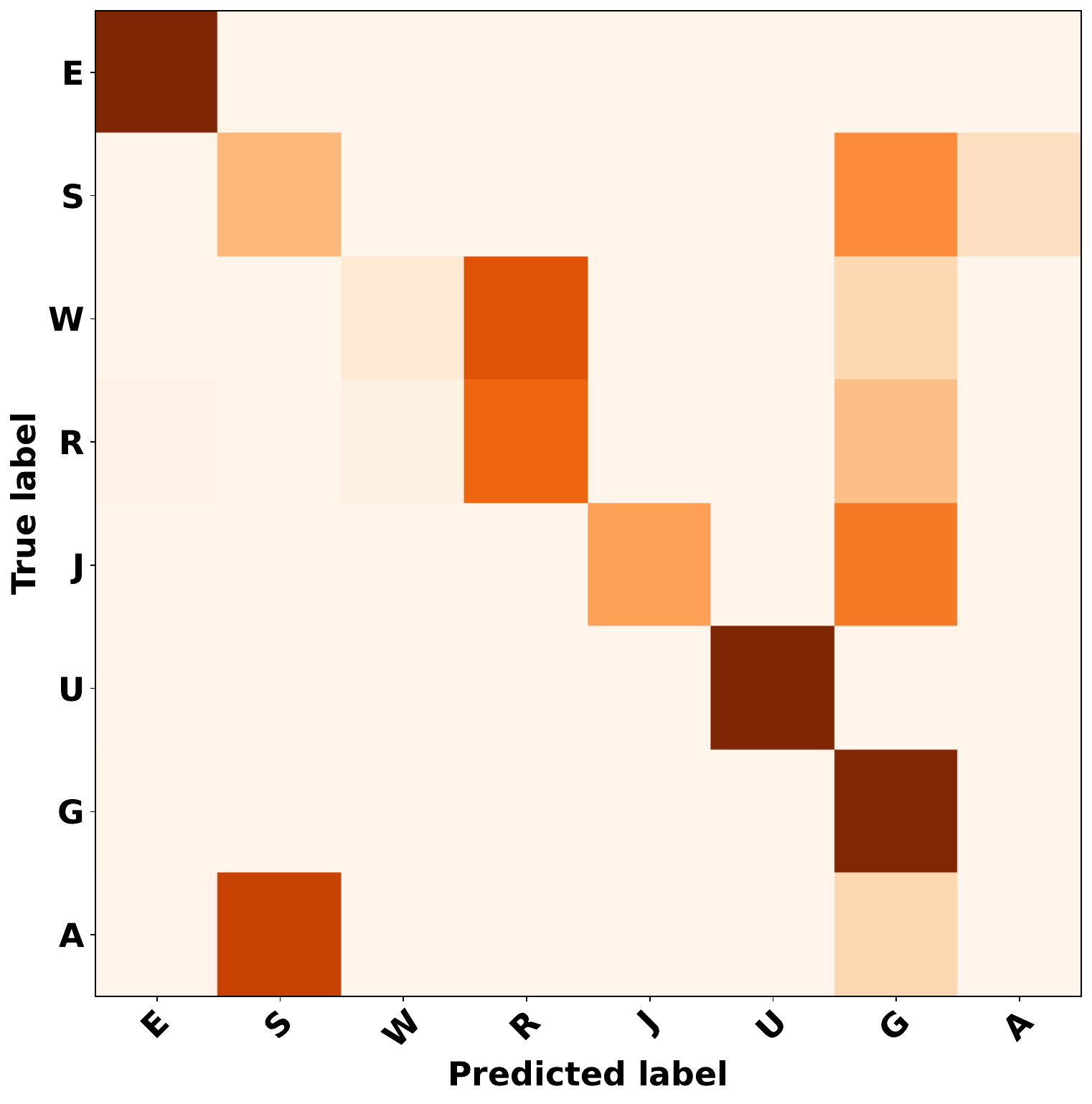}
        \text{(a) Inception - 38.38\%}
        \label{fig:conf_8_old}
    \end{minipage}
    \hfill
    \begin{minipage}[b]{0.24\textwidth}
        \centering
        \includegraphics[width=\textwidth]{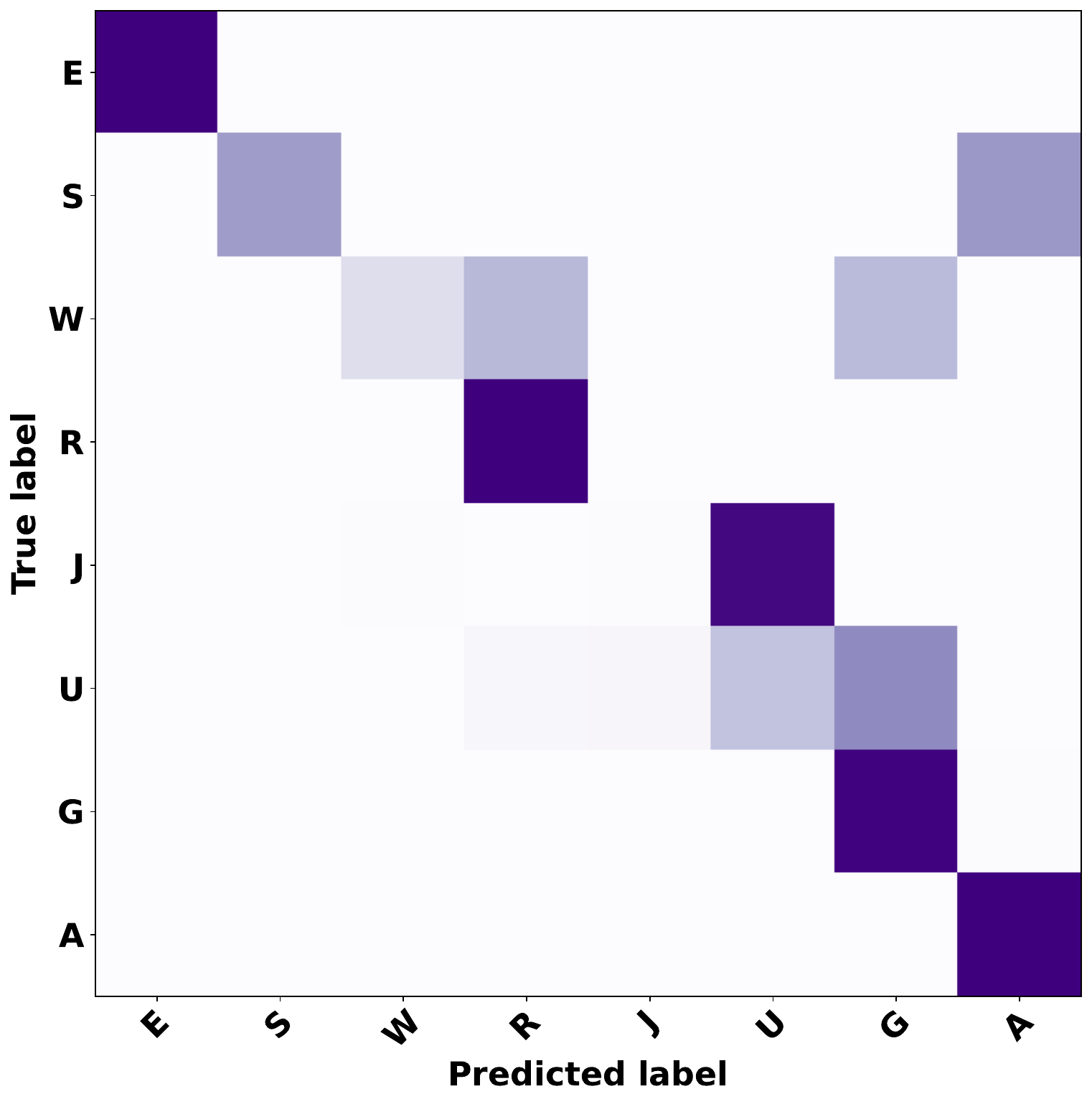}
        \text{(b) IBIS - 70.25\%}
        \label{fig:conf_8}
    \end{minipage}
     \footnotesize \textit{Note: E - Empty, S - Sitting, W - Walking, R - Running, J - Jumping, U - Stand Up, G - Standing and A - Arm Gym}
    \caption{Comparative analysis of the confusion matrix for eight movements: original Inception versus IBIS}
    \label{fig:conf_gen8}
\end{figure}

Figure \ref{fig:conf_gen8} shows the confusion matrices for eight movements, including standing, stand up, and arm gym. In Figure \ref{fig:conf_gen8}a, the performance of the original algorithm deteriorated compared to the five-movement scenario, exhibiting a higher misclassification rate for gestures such as arm gym, jumping, and sitting, resulting in an accuracy of 38.38\%. This is likely due to the SHARP algorithm's sanitization process, where the error probability increases proportionally to the number of classified gestures and bandwidth~\cite{Cominelli}. This behavior indicates that the original model struggled with overfitting and failed to generalize when more movements were added.
In contrast, Figure \ref{fig:conf_gen8}b shows that the overall accuracy remained high despite the addition of new movements, reaching 70.25\%. This superior performance is attributed to the combination of the Inception-BiLSTM hybrid  network and the SVM algorithm. Even after attempting to improve the original model's performance by modifying the dataset split to 80\% for training, 10\% for testing, and 10\% for validation, the results remained unchanged.

\subsection{Doppler Traces}

In Figure \ref{fig:neural_v3}, we present the Doppler trace plots for eight activities. It is possible to observe that the traces corresponding to the Empty, Sitting, Standing, and Arm Gym activities exhibit very similar movement patterns, making them difficult for a neural network to distinguish. Consequently, their spatial features lie very close to one another. In this context, the BiLSTM network combined with the SVM becomes the decisive component: the BiLSTM is able to extract temporal dependencies, and once these dependencies are captured, the SVM refines the decision boundaries through appropriate tuning of the trade-off parameter, kernel choice, and margin.

It becomes clear that, even with IBIS, the framework struggles to reliably distinguish highly similar Doppler traces due to the strong overlap between activities. This reinforces the need for a carefully designed preprocessing pipeline. Moreover, depending on the normalization strategy, relevant offsets and peak-related features may be suppressed, which are essential for accurate neural network inference. Nevertheless, IBIS still outperforms the original Inception network, yielding an improvement of approximately 45.30\%.

\begin{figure*}[ht!]
    \begin{center}
      \includegraphics[width=0.8\textwidth]{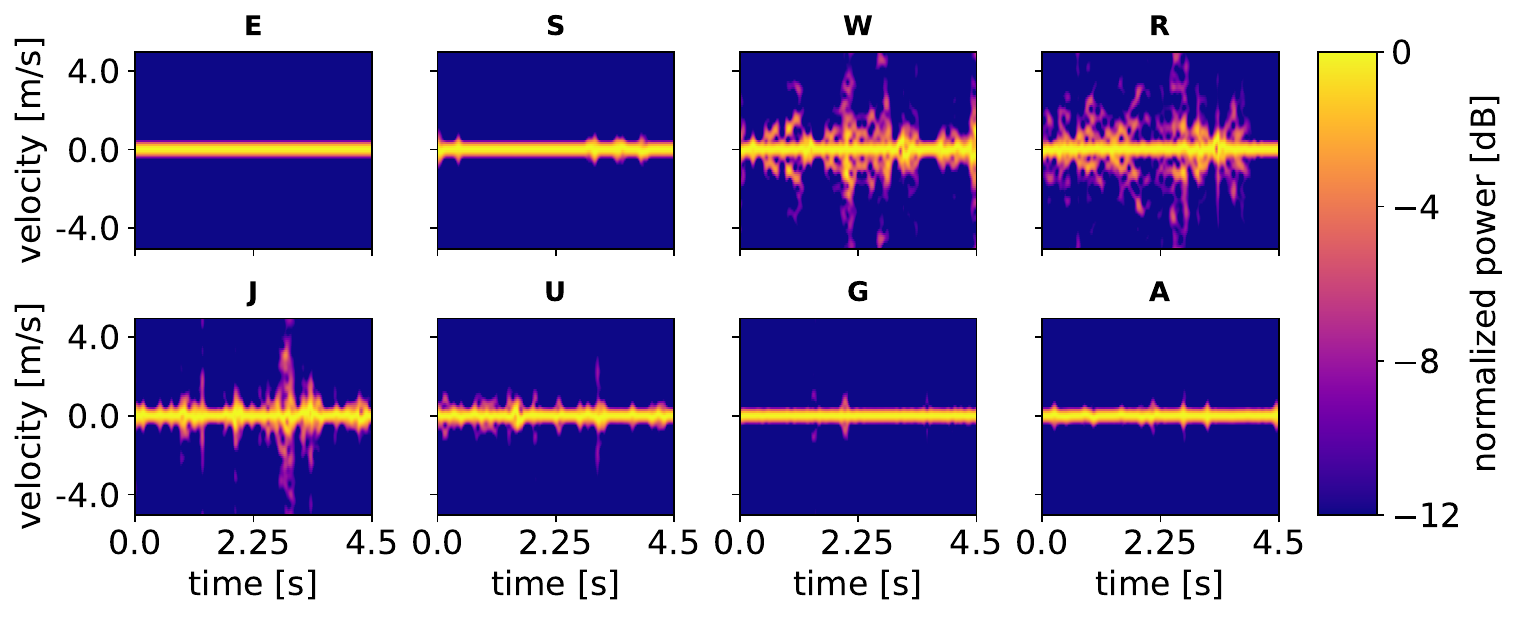}
        \caption{Doppler traces for the activities: E – Empty,
S – Sitting, W – Walking, R – Running, J – Jumping, U – Stand Up, G – Standing, A – Arm Gym.}
        \label{fig:neural_v3}
    \end{center}    
\end{figure*}

\subsection{Ablation}

\begin{figure}[hbt!]
    \begin{center}
      \includegraphics[width=0.45\textwidth]{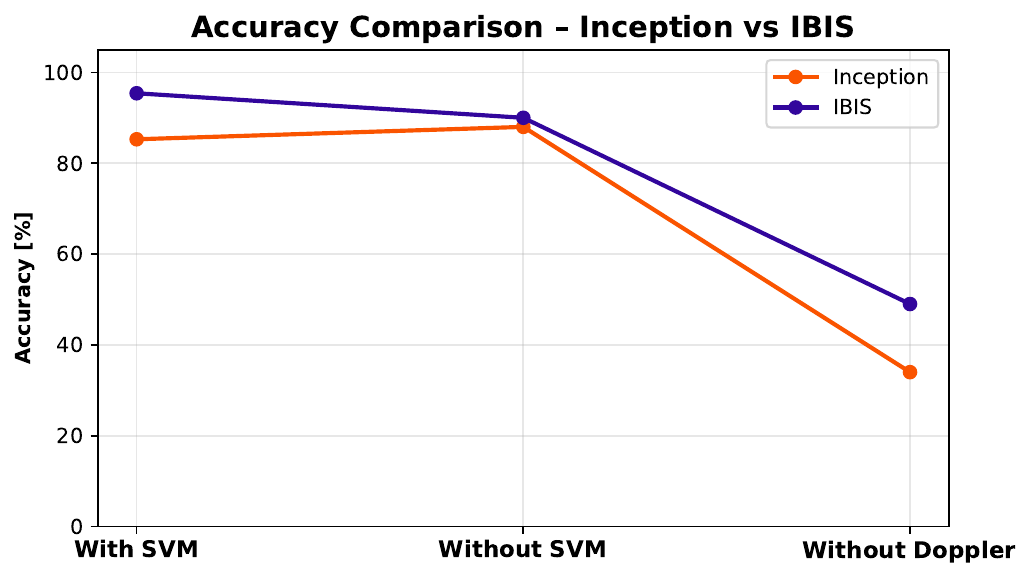}
        \caption{Graph illustrating the behavior of the accuracy when removing the Doppler traces and/or the SVM components}
        \label{fig:ablan}
    \end{center}    
\end{figure}

The next experiment conducted with our framework aims to evaluate its behavior when individual components of the pipeline are removed. First, we run the experiment using the SVM together with the Inception network and the full IBIS model. In the second step, we remove the SVM from the composition, leaving only the Inception–BiLSTM neural network. Analyzing the first experiment in Figure \ref{fig:ablan}, we observe that the Inception model with SVM experiences a reduction in performance, achieving 85.27\%, whereas the full IBIS model reaches 95.40\%.

We evaluate each component by removing it from the pipeline. Plain Inception reaches 88\% accuracy; adding the BiLSTM raises this to 90\% (a 2\% gain from temporal modeling). Adding the SVM on top of the Inception–BiLSTM backbone yields the full IBIS model at 95.40\%, the single largest jump, confirming that the post-processing SVM, not the recurrent layer alone, drives most of the gain. For contrast, attaching the SVM to Inception without the BiLSTM achieves only 85.27\%, showing that the SVM's benefit depends on the high-quality, separable features the BiLSTM provides. Removing the Doppler traces and applying IBIS to the sanitized data alone collapses performance (Inception 34\%, IBIS 49\%), confirming that Doppler phase information is essential.

Another experiment involved removing the Doppler traces from the pipeline and applying IBIS directly to the sanitized data. Analyzing Figure \ref{fig:ablan}, on the right-hand side, we observe that the Inception network without Doppler information reaches 34\% accuracy, while IBIS achieves 49\%. Although both results are lower, this confirms that Doppler features play a fundamental role in the analysis of phase components. For this specific dataset, removing Doppler information leads to a substantial degradation in accuracy, especially for the Inception network.

\subsection{Exploring another dataset}

With the intention of demonstrating the efficiency of the IBIS framework, new experiments were conducted using the public dataset provided by Cominelli et al. \cite{Cominelli}.
In this experiment, Scenario S2 was used, and—similarly to the previous experiments—the Grid Search technique was applied to select the trade-off parameter $C$ and the $\gamma$ parameter, as well as to evaluate different kernels, including RBF, Sigmoid, and Polynomial. Thus, the exact same procedure adopted in our framework was applied here, but now with a different dataset.

\begin{figure}[ht!]
    \centering
    \begin{minipage}[b]{0.24\textwidth}
        \centering
        \includegraphics[width=\textwidth]{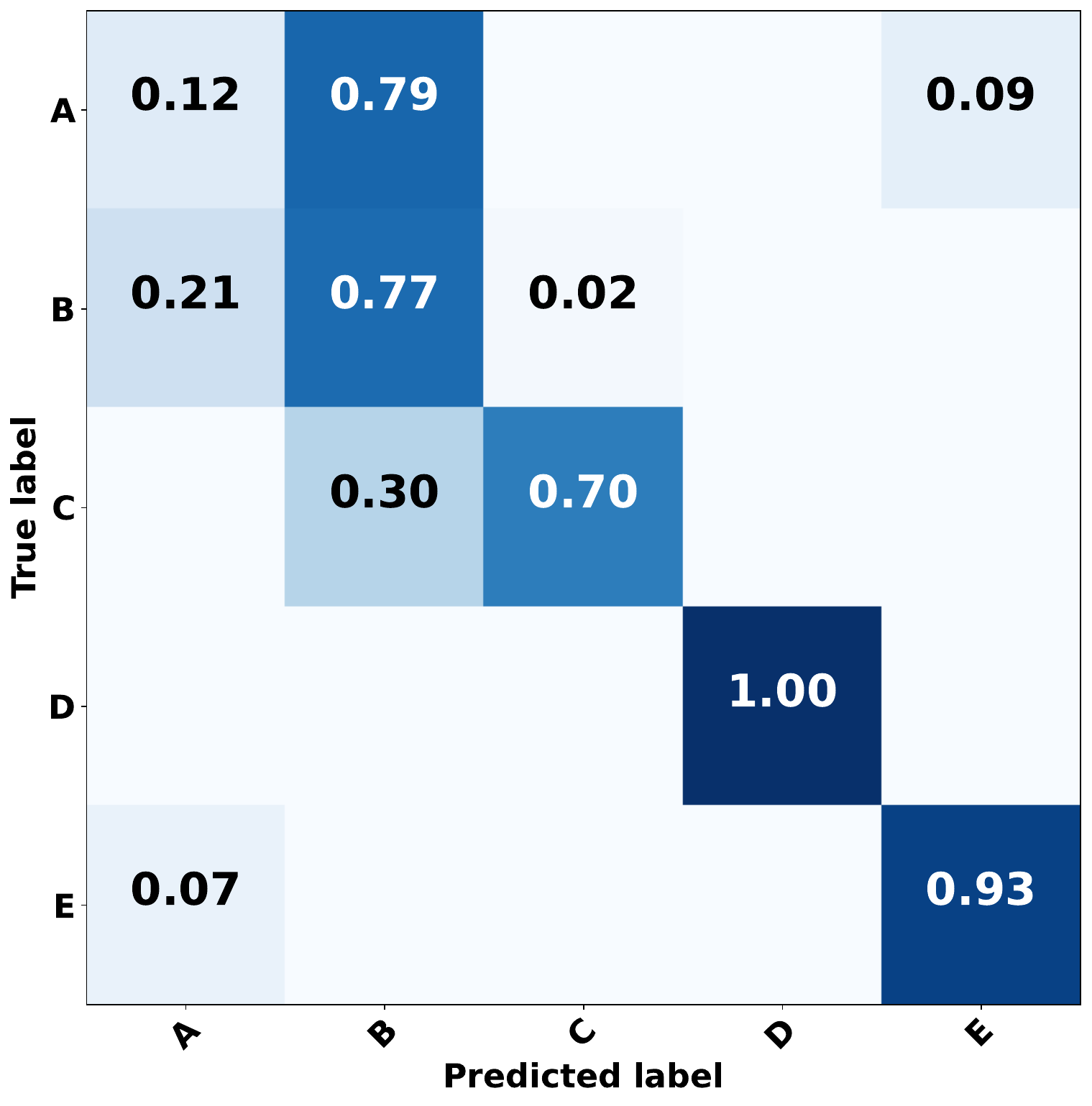}
        \text{(a) Inception - 70.23\%}
        \label{fig:conf_5_old_comm}
    \end{minipage}
    \hfill
    \begin{minipage}[b]{0.24\textwidth}
        \centering
        \includegraphics[width=\textwidth]{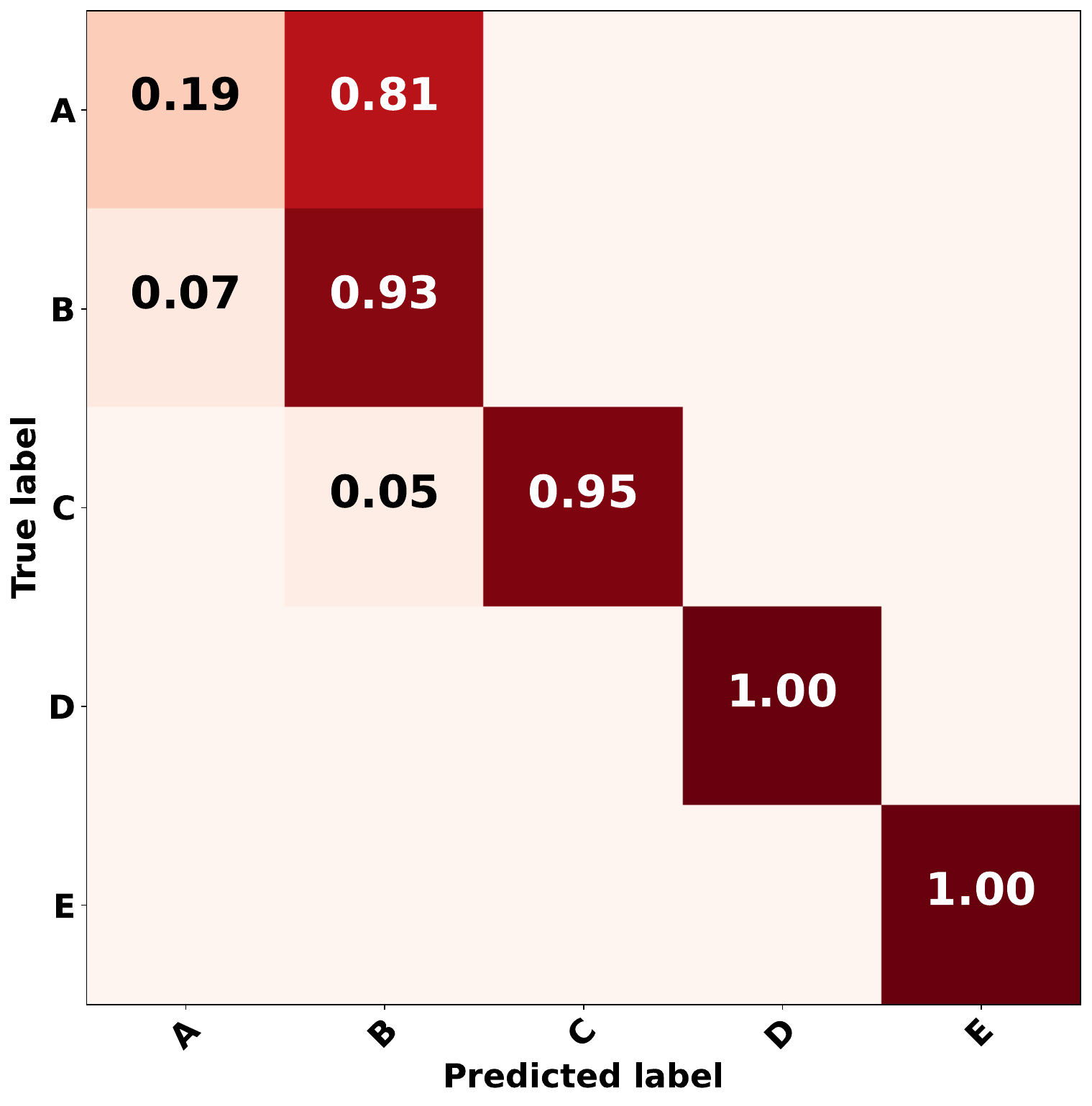}
        \text{(b) IBIS - 81.40\%}
        \label{fig:conf_5_mx_comm}
    \end{minipage}
   \hfill
         \footnotesize \textit{Note: A - Walking, B - Run, C - Jump, D - Sitting, and E - Empty Room}
    \caption{Comparative analysis of the confusion matrices for five activities: original Inception model versus the IBIS, using the dataset from Cominelli et al.~\cite{Cominelli}}
    \label{fig:conf_comm}
    \end{figure}

In this previously described dataset, several SHARP adaptations were applied to support operation with a 160 MHz bandwidth. Five activities—Walking, Running, Jumping, Sitting, and Empty 
Room, were evaluated, as shown in the confusion matrix presented in Figure \ref{fig:conf_comm}. Based on this dataset, adjustments were also made to the scripts to make them more consistent with those used in the other dataset.

It is important to highlight that, unlike the previous dataset, the IBIS framework was not able to effectively distinguish between the walking and running activities in this dataset. This limitation is mainly due to the reduced amount of available information: after generating the Doppler traces, the window segments created during the dataset splitting and training stages become significantly smaller. As a result, the IBIS does not receive sufficient information to reliably differentiate between these two activities.

Analyzing Figure \ref{fig:conf_comm}, we observe that the Inception network alone achieved an accuracy of 70.23\%, whereas the IBIS framework reached 81.40\%, representing an accuracy gain of 13.72\%. This difference in performance between datasets can be attributed to variations in data distribution; elements such as obstacles, walls, and interference introduce additional complexity and reduce recognition performance due to the increased similarity between activity patterns.

These experiments demonstrate that IBIS maintains superior performance across different datasets, achieving substantial improvements even under varying environmental conditions and inherent limitations.

\subsection{Comparable Algorithms}

\normalcolor Table \ref{tab:move} demonstrates that the IBIS framework achieved the best and most consistent results compared to other ensemble learning models. Its superior performance is attributed to the neural network's ability to capture complex patterns and the SVM's capacity to handle temporal sequence boundaries, both of which are essential for Doppler-based data. The original Inception model, for example, struggled with classifying similar movements like walking and jumping, leading to false positives and poor generalization.

\begin{table}[hbt!]
\caption{Models Accuracy}
\label{tab:move}
\begin{center}
\begin{tabular}{p{1.6cm}p{0.8cm}p{0.8cm}p{1cm}p{1cm}p{1cm}p{1cm}}\toprule
Models & Empty(\%) & Sitting(\%) & Walking(\%) & Running(\%) & Jumping(\%)   
\\ \midrule 
\textbf{IBIS}               & \textbf{100} & \textbf{87} & \textbf{91} & \textbf{100} & \textbf{99}  \\
Inception                  & 100 & 99 & 71 & 100 & 70 \\

 \bottomrule
\end{tabular}
\end{center}

\end{table}

Similarly, Table \ref{tab:average}, which presents average metrics for five movement classes, confirms that our IBIS outperformed other tested models. We included additional metrics like F1-score, recall, and precision to provide a more comprehensive evaluation, as relying solely on accuracy can be misleading. These metrics help assess the model's performance in imbalanced scenarios and its ability to balance precision and recall.

\begin{table}[hbt!]
\caption{Average Models Metrics for 5 Movements}
\label{tab:average}
\begin{center}
\begin{tabular}{p{1.7cm}p{1.2cm}p{1.5cm}p{1.0cm}p{1.2cm}}\toprule
Models & Accuracy(\%) & F1 Score(\%) & Recall(\%) & Precision(\%)   
\\ \midrule 
\textbf{IBIS}               & \textbf{95.40} & \textbf{95.21} & \textbf{95.30} & \textbf{95.57}  \\
Inception                  & 88.16 & 88.26 & 87.87 & 91.47  \\

 \bottomrule
\end{tabular}

\end{center}
\end{table}

While the CNN-ABLSTM model from a related study~\cite{He} presents a lower computational cost due to its lighter architecture, it achieves only 85.82\% accuracy, whereas our IBIS method reaches 95.40\%. This 10.04\% improvement under the same 100-epoch training condition is primarily due to the limited number of attention layers and the reduced architectural complexity in their model, which hinder its ability to capture complex temporal–spatial patterns and subtle Doppler-induced variations. In contrast, our IBIS framework leverages the Inception module for spatial feature extraction and the BiLSTM module for temporal dependency learning, making it more robust to noise, interference, and reflective-surface effects. Additionally, our pipeline integrates advanced signal processing, the SHARP algorithm for noise suppression, and traditional machine learning SVM for refined classification. This combined approach significantly improves generalization compared to the transfer-learning-based model trained on raw, unprocessed data in ~\cite{He}.

\section{Comparison Between Softmax and SVM Clustering Representations}

\begin{figure*}[ht!]
    \centering
    \makebox[\textwidth][c]{%
       \includegraphics[width = 1.0\textwidth]{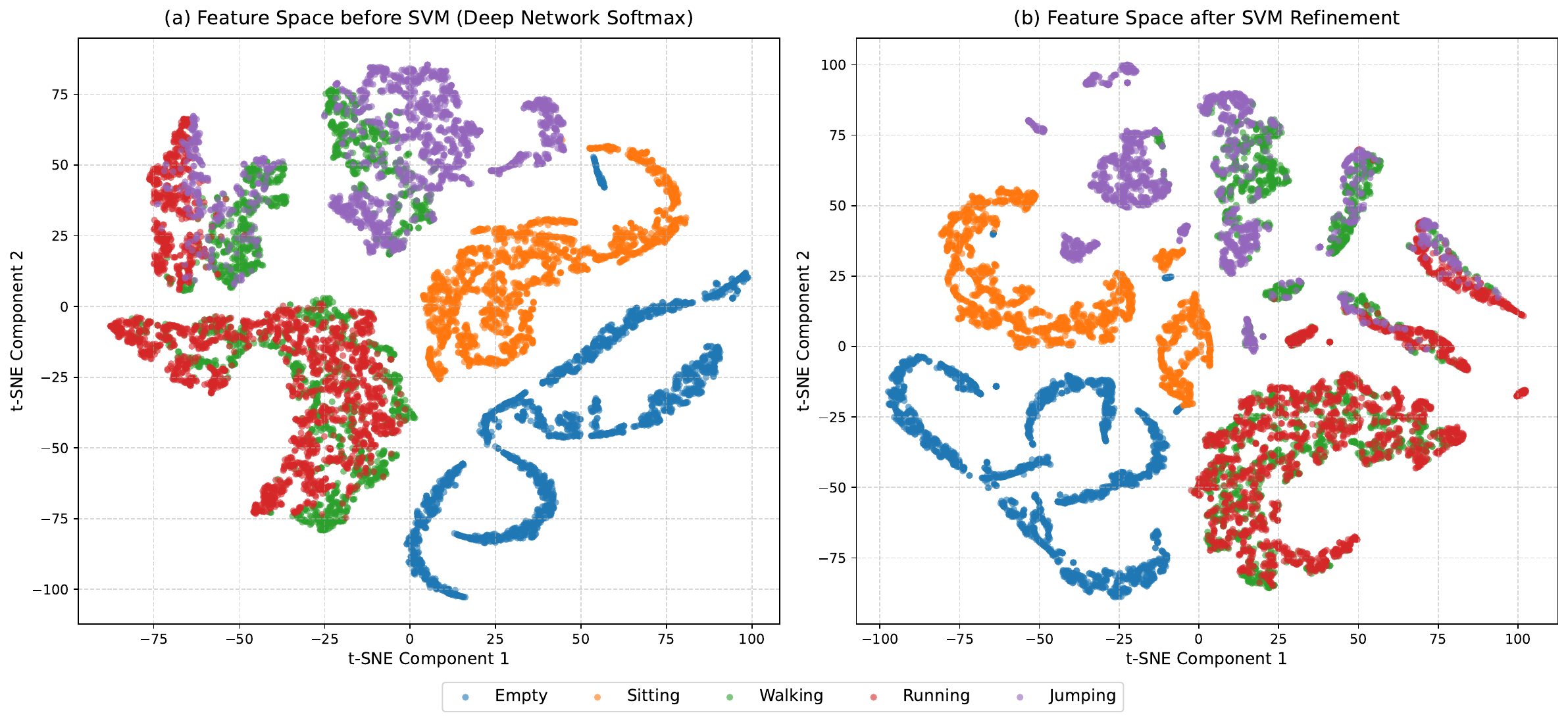}
    }
    \hfill
    \begin{minipage}[b]{0.48\textwidth}
        \begin{center}
          \text{\small (a)}
        \end{center}
          \label{fig:svm_a}
    \end{minipage}
    \hfill
    \begin{minipage}[b]{0.48\textwidth}
        \begin{center}
       \text{\small (b)}
        \end{center}
        \label{fig:svm_b}
    \end{minipage}
    \caption{A Comparative Analysis of Softmax and SVM Classifiers Using t-SNE for Class Separation and Clustering Representation}
    \label{fig:svm_}
\end{figure*}

To demonstrate the importance of incorporating the SVM into the pipeline, we generated a t-SNE \cite{Chen} visualization that explicitly compares the class distributions obtained from the softmax classification against those from the SVM-based approach, as illustrated in Figure \ref{fig:svm_}.

First, analyzing the softmax scenario in Figure \ref{fig:svm_}a, convergence is observed where the clusters appear distinct, separating the classes reasonably well. Classes such as "Empty" and "Sitting" form clusters in specific regions of the plot. However, a major issue can be observed within the remaining classes, which exhibit significant overlap in the lower and central regions of the graph. Consequently, if the softmax layer relied on a single decision boundary, the model would yield a considerable rate of false positives.

Conversely, analyzing Figure \ref{fig:svm_}b, the data points are distributed in elongated bands with more compact and dense clusters, aligns with the SVM objective of finding a maximum-margin hyperplane. The most critical improvement occurred in the Walking and Running classes, where the overlap decreased drastically, making the cluster boundaries much sharper and more defined. Furthermore, the Jumping class became completely isolated. Thus, the SVM successfully captured class nuances without significant overlap.

It is worth noting that while perfect class separation can sometimes indicate an overfitted network leading to ambiguity, the SVM mitigates this by maximizing the margin between boundaries, effectively correcting the class overlaps that occur when utilizing the softmax layer alone.

\subsection{Performance Comparison Among Multiple Deep Learning Algorithms
}

Table \ref{tab:average_models} presents a comparison of accuracy, precision, recall, and F1-score across different deep learning model architectures. It is worth noting that the core pipeline remains intact; only the deep learning network was altered, while the post-processing stage using the SVM remains under the exact same specifications.

\begin{table*}[hbt!]
\caption{Performance Comparison and Impact of Alternative Deep Learning Architectures}
\label{tab:average_models}
\begin{center}
\setlength{\tabcolsep}{18.5pt} 
\begin{tabular}{lccccc}
\toprule
\textbf{Architecture} & \textbf{\shortstack{Accuracy(\%)}} & \textbf{\shortstack{F1 Score(\%)}} & \textbf{\shortstack{Recall(\%)}} & \textbf{\shortstack{Precision(\%)}} & \textbf{\shortstack{Impact Acc($\Delta\%$) }} \\ \midrule 
\textbf{IBIS (Proposed)} & \textbf{95.40} & \textbf{95.21} & \textbf{95.30} & \textbf{95.57} & \textbf{Reference} \\ \midrule
Inception-BiGRU          & 91.12          & 90.45          & 90.01          & 93.93          & $-4.28$            \\
Inception-TCN            & 92.93          & 92.40          & 92.05          & 94.26          & $-2.47$            \\
Inception-Transformer    & 80.04          & 79.51          & 79.97          & 85.51          & $-15.36$           \\
ResNet-BiLSTM            & 80.84          & 78.07          & 78.68          & 80.48          & $-14.15$           \\
\bottomrule
\end{tabular}
\end{center}
\end{table*}

Initially, we introduce the results obtained after 10 simulations, from which the average was calculated for inclusion in the table. Under this framework, we introduced the Inception-BiGRU, a variation of the Inception-BiLSTM where a Gated Recurrent Unit (GRU) was incorporated to process spatial and temporal sequences. This modification reduces memory consumption and accelerates training times. However, as observed in the presented results, there was an accuracy degradation on the order of 4.28\%. The explanation behind this behavior is that without the convolutional signaling inherent to the BiLSTM architecture, the Doppler signal loses precision when attempting to correlate the end with the beginning of the sequence. Furthermore, regarding sequence length, longer sequences increase the risk of generating false positives.

We also evaluated a hybrid Inception-TCN architecture, which replaces the BiLSTM recurrent blocks with a Temporal Convolutional Network (TCN). The TCN processes the entire sequence in parallel using convolutional blocks driven by dilated convolutions. Consequently, training becomes faster and memory usage is reduced. Nevertheless, this architectural shift introduced an impact, resulting in an accuracy drop of approximately 2.47\%. This reduction was triggered by the vanishing gradient problem, where very long sequences suffer from successive multiplications, alongside a severe risk of overfitting, where the model memorizes the training data rather than generalizing the underlying features.

Additionally, a hybrid Inception-Transformer structure was evaluated with an implicit constraint. Due to the excessive number of parameters and the high computational cost regarding VRAM, we froze all internal layers of the Transformer network and left only the classification layer trainable. Although this strategy reduced the model's computational weight, it resulted in an accuracy drop on the order of 15.36\%. This abrupt performance decline is justified by the inherent freezing of the hidden layers and the utilization of only a single attention layer, whereas achieving optimal Transformer performance typically requires a multi-head attention mechanism.

Finally, we compared our architecture against a ResNet-BiLSTM framework. In this setup, the BiLSTM component was kept exactly identical, but the Inception network was removed and replaced by a ResNet. This modification led to a performance reduction on the order of 14.15\%. By removing the Inception module, the model loses its multiscale feature extraction capability; the Inception block processes features at multiple scales simultaneously, unlike ResNet, which processes them sequentially. While ResNet is highly recommended for deeper networks, it can be suboptimal for feature extraction in shallower configurations.

\subsection{Computational Complexity}

\begin{figure}[hbt!]
    \begin{center}
      \includegraphics[width=0.45\textwidth]{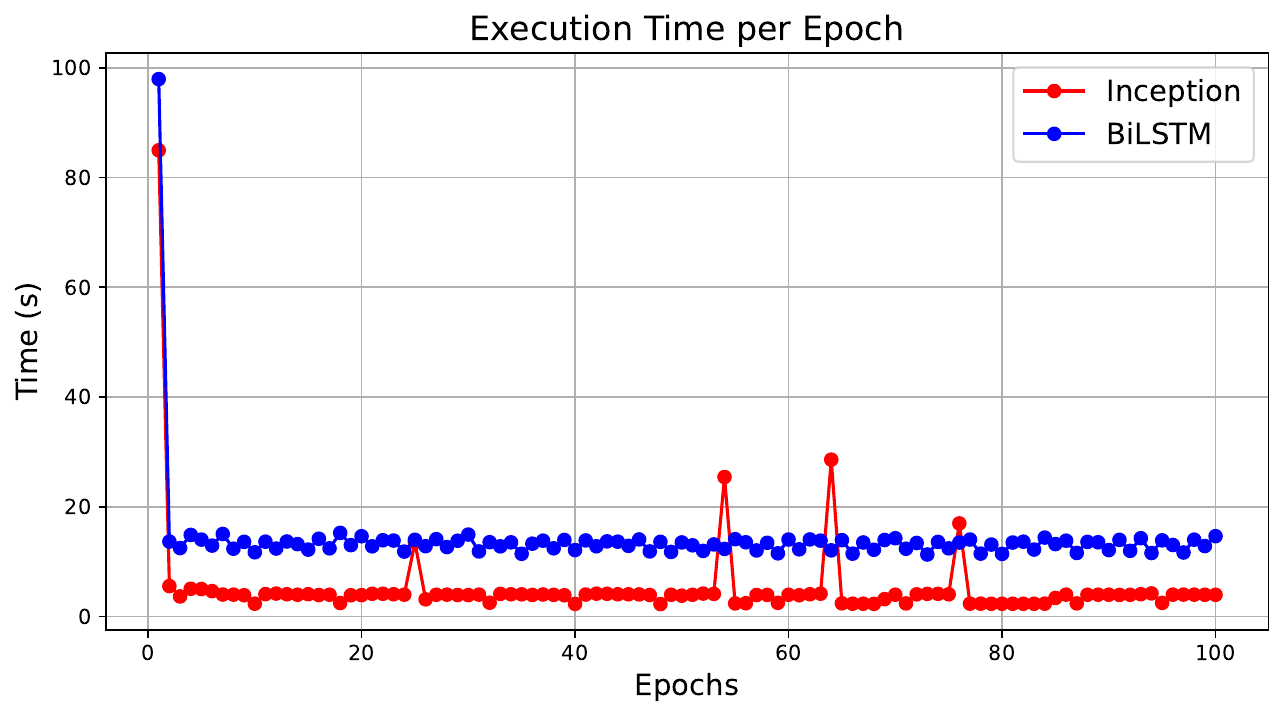}
        \caption{Graph illustrating the execution time per training epoch for both the original Inception architecture and the proposed hybrid IBIS model.}
        \label{fig:comp_complex}
    \end{center}    
\end{figure}

A primary challenge with neural networks is computational cost, as a deeper network with more layers typically consumes more resources. To objectively measure this, Figure \ref{fig:comp_complex} shows the computational cost over 100 training epochs without using callbacks like early stopping.

We evaluated two distinct architectures:

\begin{itemize}
    \item The original Inception network;
    \item An Inception network combined with a BiLSTM layer.
\end{itemize}

The standalone Inception network had the lowest computational cost and the shortest training time per epoch due to its shallow architecture. However, it failed to accurately classify movements, producing false positives and showing temporal instability.
It should be noted that the SVM classifier is not displayed in this figure because it is not involved in the epoch-level training process. The SVM is fitted only once after the neural network has converged, using the extracted feature probabilities. Consequently, its computational overhead is negligible, and its inclusion in the graphical representation is unnecessary.

\section{Conclusion}
\label{sec:conclusao}

This paper addresses the challenges of generalization and robustness in HAR using non-intrusive Wi-Fi sensing. We propose IBIS, a novel hybrid architecture that strategically integrates Inception layers for spatial feature extraction, a BiLSTM module for temporal modeling, and an SVM classifier for sharper decision boundaries. Extensive evaluations across scenarios with different activity sets (five and eight classes) demonstrate the framework’s effectiveness, reaching an average accuracy of 95.40\%. Notably, IBIS outperforms the state-of-the-art SHARP method by 7.58\% on Dataset A, highlighting its superior capability in handling noise and environmental interference.
A second improvement of 13.40\% was observed on Dataset B, largely due to environmental differences, where variations in surface materials and layout can alter signal propagation patterns. Nevertheless, even under new conditions and datasets, IBIS consistently delivers superior performance. The analysis further confirms that replacing the standard Softmax layer with an SVM classifier significantly enhances classification performance in complex environments. Future work will focus on optimizing the Doppler trace acquisition and pre-processing stages to reduce computational complexity and further streamline the training pipeline. Additionally, we suggest utilizing alternative hardware capture models that are not based on Nexmon, such as the Atheros CSI Tool, to evaluate cross-domain generalization across different CSI (Channel State Information) data extraction frameworks. 
Moreover, we will investigate motion dynamics features incorporating signal delay, complemented by advanced deep learning algorithms; this synchronization will enable the extraction of more precise information.

\bibliographystyle{IEEEtran} 
\bibliography{references}

\begin{IEEEbiography}[{\includegraphics[width=1in,height=1.25in,clip,keepaspectratio]{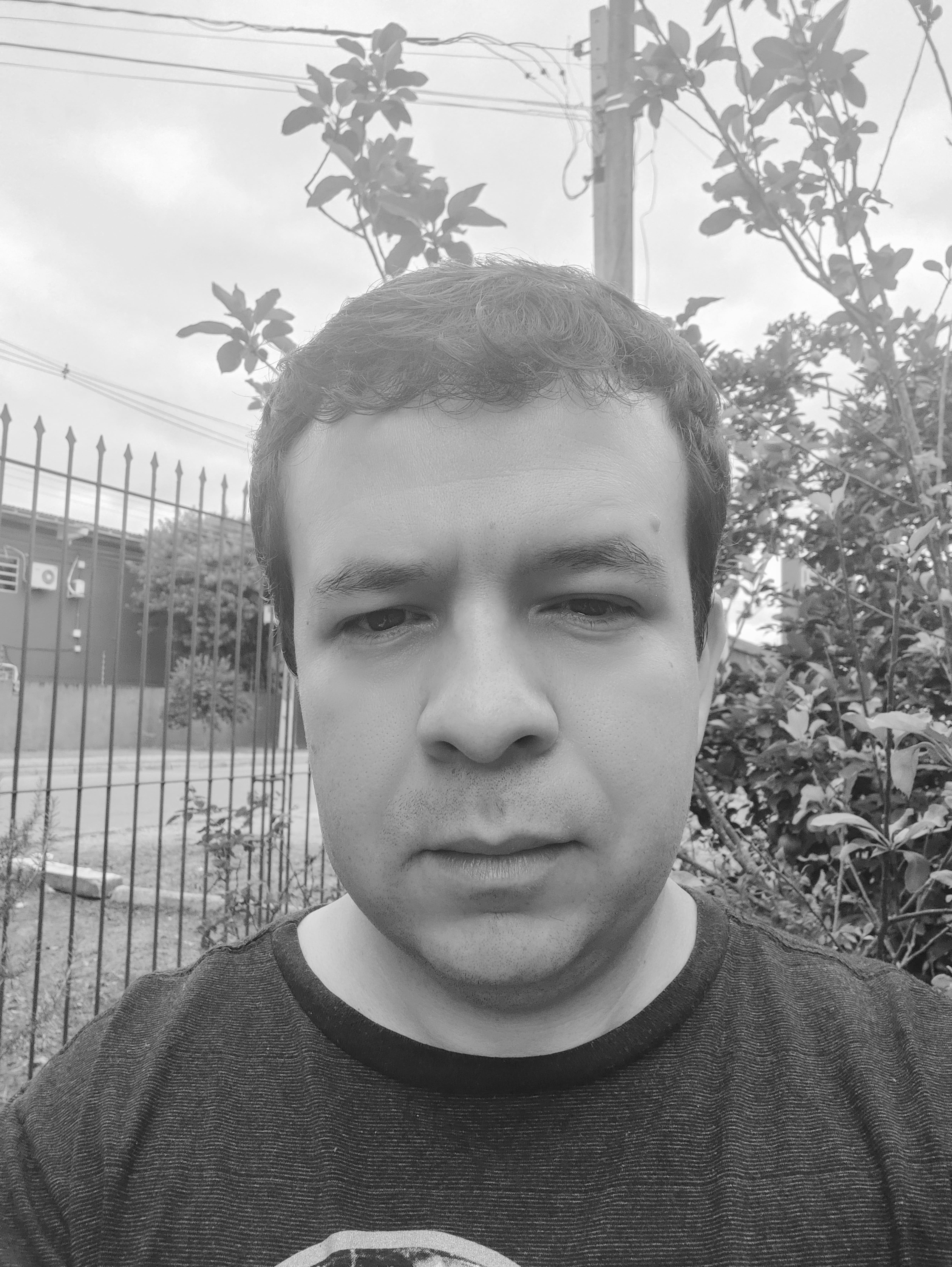}}]{ALISON M. FERNANDES}~received a B.Sc. degree in Electrical Engineering from the Federal University of Technology – Paraná (UTFPR) in 2016. In 2021, he completed a second B.Sc. degree in Electronic Engineering at the same university. In 2025, he received his M.Sc. degree. He is currently pursuing a Ph.D. in Electrical Engineering and Industrial Informatics at UTFPR. During his undergraduate studies, he conducted research in power systems, electronics, and embedded systems. His main areas of interest include programming, intelligent networks, machine learning, artificial neural networks, and generative artificial intelligence (AI) networks.

\end{IEEEbiography}
\begin{IEEEbiography}[{\includegraphics[width=1in,height=1.25in,clip,keepaspectratio]{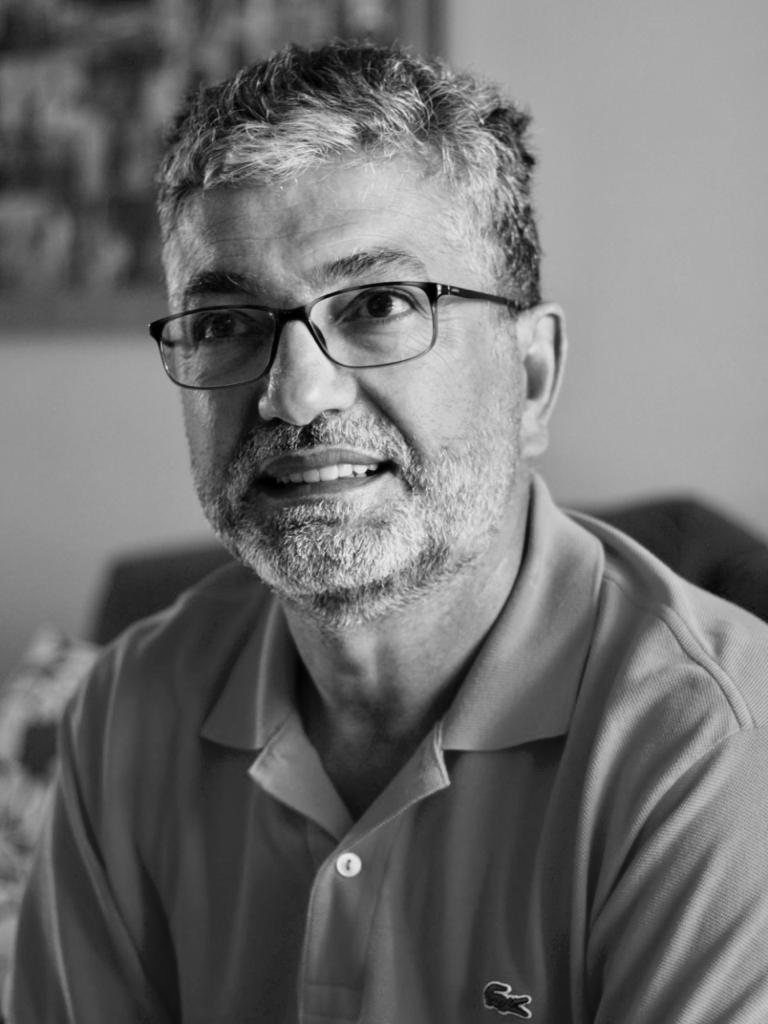}}]{HERMES I. DEL MONEGO}~received the B.Sc. Degree in Computer Science from Positivo University Brazil in 2000, an M.Sc. degree from the Federal University of Technology - Paraná (UTFPR), and a Ph.D. in Electrical Engineering and Computing from the Faculty of Engineering of Porto University (FEUP) in 2011. Since 2010, he has been with the Electronics Department at the Federal University of Technology – Paraná (UTFPR), where he is now an associate professor. He has been invited to conduct research at INESC TEC, Porto, Portugal, and is the academic director at INESC P\&D Brazil. His main research topics include computer networks, cellular networks, radio resource management, traffic distribution, and handover management.
  
\end{IEEEbiography}
\begin{IEEEbiography}[{\includegraphics[width=1in,height=1.25in,clip,keepaspectratio]{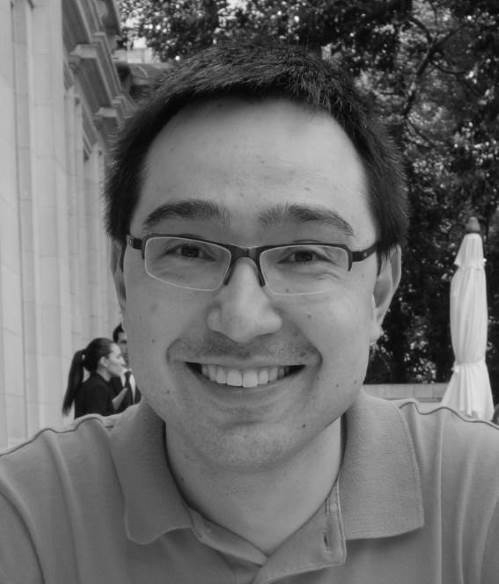}}]{BRUNO S. CHANG}~was born in Curitiba, Paraná, Brazil, in 1984. He received the B.Sc. Degree in Telecommunications Engineering from the Regional University of Blumenau (FURB), Brazil, in 2006, the M.Sc. degree in electrical engineering from the Federal University of Santa Catarina (UFSC), Florianópolis, Brazil, in 2008, and the D.Sc. degree in electrical engineering from the Federal University of Santa Catarina (UFSC), Florianópolis, Brazil, and the Conservatoire National des Arts et Métiers (CNAM), Paris, France in 2012. He was a visiting professor at CNAM in 2018 and a visiting researcher at INESC TEC Porto, Portugal in 2024. He is currently an Associate Professor at the Federal University of Technology - Paraná (UTFPR), Curitiba, Brazil. His main research interests lie in the areas of digital communications and signal processing, including multi-carrier systems, detection and estimation algorithms, widely linear processing and energy efficient communications.  
\end{IEEEbiography}

\begin{IEEEbiography}
[{\includegraphics[width=1in,height=1.25in,clip,keepaspectratio]{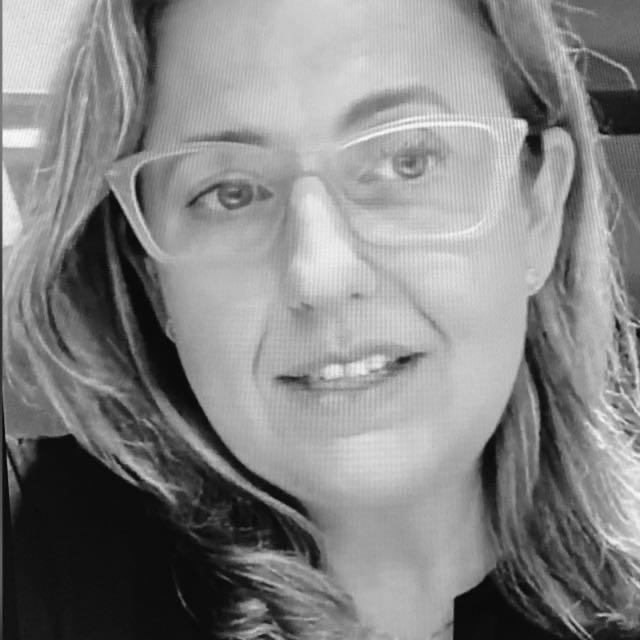}}]{ANELISE MUNARETTO}~received the Dipl.Ing. Degree in Computer Engineering from the Pontifical Catholic University of Parana, Brazil, in 1994, and the M.S. and Ph.D. degrees in Computer Science from Université Pierre et Marie Curie (UPMC–Sorbonne Universités), in 2001 and 2004, respectively. Since 2005, she has been with the Federal University of Technology–Paraná (UTFPR), where she is currently a Full Professor. Her research interests include computer networks with an emphasis on domains related to mobile networking, the Internet of Things (IoT), and smart cities.
\end{IEEEbiography}

\begin{IEEEbiography}
[{\includegraphics[width=1in,height=1.25in,clip,keepaspectratio]{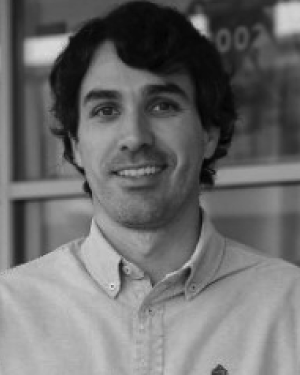}}]{RUI CAMPOS}~received
the Ph.D. degree in electrical and computer engineering from the University of Porto, in 2011. He is currently a Senior Researcher and a Coordinator of the Centre for Telecommunications and Multimedia, INESC TEC, and an Assistant Professor with the University of Porto, where he teaches courses on telecommunications. He has
coordinated several research projects, including the WiFIX project in CONFINE Open Call 1,
SIMBED (Fed4FIRE+OC3), and SIMBED+ (Fed4FIRE+OC5). He has also participated in several EU research projects, including H2020 ResponDrone, H2020 RAWFIE, FP7 SUNNY, and FP6 Ambient Networks. He is the author
of more than 85 scientific publications in international conferences and journals with peer review. His research interests include airborne, maritime, and green networks, with a special focus on medium access control, radio resource management, and network auto configuration. He has been a TPC Member of several international conferences, including IEEE INFOCOM, IEEE ISCC, and IFIP/IEEE Wireless Days. He was the General Chair of Wireless Days 2017 and the TPC Vice-Chair of 2021 Joint EuCNC and 6G.
Summit.
\end{IEEEbiography}

\begin{IEEEbiography}
[{\includegraphics[width=1in,height=1.25in,clip,keepaspectratio]{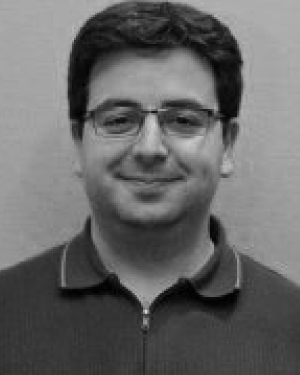}}]{HELDER FONTES}~received the M.Sc. degree in 2010 and the Ph.D. degree in 2019, in informatics engineering from the Faculty of Engineering,
University of Porto, Portugal. He has been an advisor of more than 10 M.Sc. theses on wireless
networking simulation, emulation, and experimentation. He is currently the Coordinator of
the Wireless Networks (WiN) area with INESC TEC, and since 2009, he has been participating
in multiple national and EU research projects, including SITMe, HiperWireless, FP7 SUNNY, H2020 ResponDrone,
DECARBONIZE, FLY.PT and Fed4FIRE+ SIMBED, and SIMBED+ and SMART open call projects. His research interests include wireless networking simulation, emulation, and experimentation in the scope of emerging scenarios such as airborne and maritime, with a special focus on repeatability and reproducibility of experiments using digital twins of wireless testbeds.
\end{IEEEbiography}

\end{document}